\documentclass[10pt,journal,compsoc]{IEEEtran}
\usepackage{etoolbox}
\usepackage{caption}
\usepackage{subcaption}
\usepackage{amsmath,amsfonts}
\usepackage{algorithmic}
\usepackage{array}
\usepackage{textcomp}
\usepackage{stfloats}
\usepackage{url}
\usepackage{verbatim}
\usepackage{graphicx}
\usepackage[colorlinks,linkcolor=red,anchorcolor=blue,citecolor=green]{hyperref}
\def\BibTeX{{\rm B\kern-.05em{\sc i\kern-.025em b}\kern-.08em
    T\kern-.1667em\lower.7ex\hbox{E}\kern-.125emX}}
\usepackage{balance}
\usepackage{color}
\usepackage{amssymb}
\usepackage{amsmath}  
\usepackage{bm}
\usepackage{wrapfig}
\usepackage{bbm}
\usepackage{mathrsfs}
\usepackage{algorithm}
\usepackage{algorithmic}
\usepackage{multirow}
\usepackage{bm}
\usepackage{bbding}
\usepackage{pifont}
\usepackage[table]{xcolor}
\newcommand{\cmark}{\ding{51}}%
\usepackage{colortbl}
\definecolor{Gray}{gray}{0.9}
\newcommand{\sssection}[1]{\noindent\textbf{#1}}
\newcolumntype{I}{!{\vrule width 1pt}}
\usepackage[super]{nth}

\makeatletter
\newcommand{\thickhline}{%
    \noalign {\ifnum 0=`}\fi \hrule height 1pt
    \futurelet \reserved@a \@xhline
}
\makeatother

\usepackage{arydshln}
\definecolor{mydarkblue}{rgb}{0,0.08,0.45}
\newcommand{\pub}[1]{\color{gray}{\tiny{[{#1}]}}}
\usepackage[utf8]{inputenc}

\usepackage{listings}

\definecolor{bblue}{RGB}{0,30,95}
\definecolor{rred}{RGB}{190,0,0}
\definecolor{mygray}{gray}{.9}
\definecolor{ggray}{RGB}{127,127,127}
\definecolor{sblue}{RGB}{0,173,206}
\definecolor{ppink}{RGB}{240,46,142}
\newcommand{\ie}{\textit{i}.\textit{e}.}
\newcommand{\eg}{\textit{e}.\textit{g}.}

\newcommand{\cf}{\textit{cf}.}
\newcolumntype{y}[1]{>{\raggedright\arraybackslash}p{#1pt}}
\newcolumntype{z}[1]{>{\raggedleft\arraybackslash}p{#1pt}}

\definecolor{patch}{RGB}{216,118,52}
\definecolor{frame}{RGB}{77,85,129}
\definecolor{proto}{RGB}{112,48,160}
\definecolor{temp}{RGB}{64,136,201}

\begin{document}
\title{Spatio-temporal Decoupled Knowledge  \\ Compensator for Few-Shot Action Recognition}
\author{Hongyu Qu, Xiangbo Shu,~\IEEEmembership{Senior Member, IEEE}, Rui Yan, Hailiang Gao, Wenguan Wang,~\IEEEmembership{Senior Member, IEEE}, Jinhui Tang,~\IEEEmembership{Senior Member, IEEE}
\thanks{
\textit{H. Qu, X. Shu, R. Yan, H. Gao, and J. Tang are with the School of Computer Science and Engineering, Nanjing University of Science and Technology, Nanjing 210094, China. E-mail: \{quhongyu, shuxb, ruiyan, gaohailiang, jinhuitang\}@njust.edu.cn. (Corresponding author: Xiangbo Shu)}
        
\textit{W. Wang is with the State Key Lab of Brain-Machine Intelligence, Zhejiang University, China. (E-mail: wenguanwang.ai@gmail.com)}
}
}
\markboth{IEEE TRANSACTIONS ON PATTERN ANALYSIS AND MACHINE INTELLIGENCE, 2026}%
{Shell \MakeLowercase{\textit{et al.}}: A Sample Article Using IEEEtran.cls for IEEE Journals}

\IEEEtitleabstractindextext{
\begin{abstract}
Few-Shot Action Recognition (FSAR) is a challenging task that requires recognizing novel action categories with a few labeled videos. Recent works typically apply semantically coarse category names as auxiliary contexts to guide the learning of discriminative visual features. However, such context provided by the action names is too limited to provide sufficient background knowledge for capturing novel spatial and temporal concepts in actions. In this paper, we propose \textbf{\textsc{DiST}}, an innovative \textbf{D}ecomposition-\textbf{i}ncorporation framework for FSAR that makes use of decoupled \textbf{S}patial and \textbf{T}emporal knowledge provided by large language models to learn expressive multi-granularity prototypes. In the decomposition stage, we decouple vanilla action names into diverse spatio-temporal attribute descriptions (action-related knowledge). Such commonsense knowledge complements semantic contexts from spatial and temporal perspectives. In the incorporation stage, we propose Spatial/Temporal Knowledge Compensators (SKC/TKC) to discover discriminative object-level and frame-level prototypes, respectively. In SKC, object-level prototypes adaptively aggregate important patch tokens under the guidance of spatial knowledge. Moreover, in TKC, frame-level prototypes utilize temporal attributes to assist in inter-frame temporal relation modeling. These learned prototypes thus provide transparency in capturing fine-grained spatial details and diverse temporal patterns. Experimental results show \textsc{DiST} achieves state-of-the-art results on five standard FSAR datasets. Our source code is available at \href{https://github.com/quhongyu/DiST}{DiST}.
\end{abstract}
\begin{IEEEkeywords}
    Few-shot Learning, Action Recognition, Prototype Learning, Decoupled Knowledge.
\end{IEEEkeywords}}

\maketitle

\IEEEdisplaynontitleabstractindextext
\IEEEpeerreviewmaketitle

\section{Introduction}
\IEEEPARstart{W}{ith} deep learning advancements~\cite{wang2016temporal,lin2019tsm,wang2023videomae,shu2019hierarchical,shu2021spatiotemporal,yan2020higcin,wang2025visual}, significant progress has been made in the field of action recognition~\cite{yang2020temporal,wang2021tdn,huang2024matching,lin2024vlg,shu2022multi,sun2022human,yan2023progressive} recently. However, this success relies heavily on a large amount of manually-labeled samples, which are time-consuming and expensive to acquire. To alleviate the data-hunger issue, considerable works~\cite{zhang2020few,cao2020few,li2022ta2n,boosting,xia2023few} have turned their attention to few-shot action recognition (FSAR), where there exist base action classes (seen) with a large volume of training examples, and novel action classes (previously unseen) with unlabeled samples. FSAR first learns feature representation from base action classes, and then evaluates its generalization ability on novel action classes.

Modern  FSAR solutions~\cite{zheng2024saliency,perrett2021temporal,zheng2022few,nguyen2022inductive,wang2023few} are largely built upon metric-based meta-learning paradigm~\cite{vinyals2016matching}, where the model learns class (prototype) representation and performs prototype-query matching with respect to predefined or learned distance metrics. Among them, top-leading methods~\cite{cao2020few,thatipelli2022spatio,wu2022motion,boosting} directly extract class-related spatio-temporal feature representation from raw visual signals. Though impressive, these methods lack a basic grasp of explicit action knowledge, struggling to learn new concepts in action classes, particularly under data-limited conditions. Recent works~\cite{wang2023clip,qu2024mvp,xu2024lvlm,huang2024noise,qu2025learning} transfer knowledge from pre-trained vision-language models~\cite{jia2021scaling,radford2021learning} (\eg, CLIP~\cite{radford2021learning}) to enhance FSAR model capability. However, these methods typically apply semantically coarse or ambiguous category names as auxiliary context information to compensate for visual features. Such context provided by the action names is too limited to provide enough background knowledge for video action understanding~\cite{ni2022expanding,wang2023actionclip,wu2024transferring}.
\begin{figure}[t]
    \centering
   \includegraphics[width=0.5\textwidth]{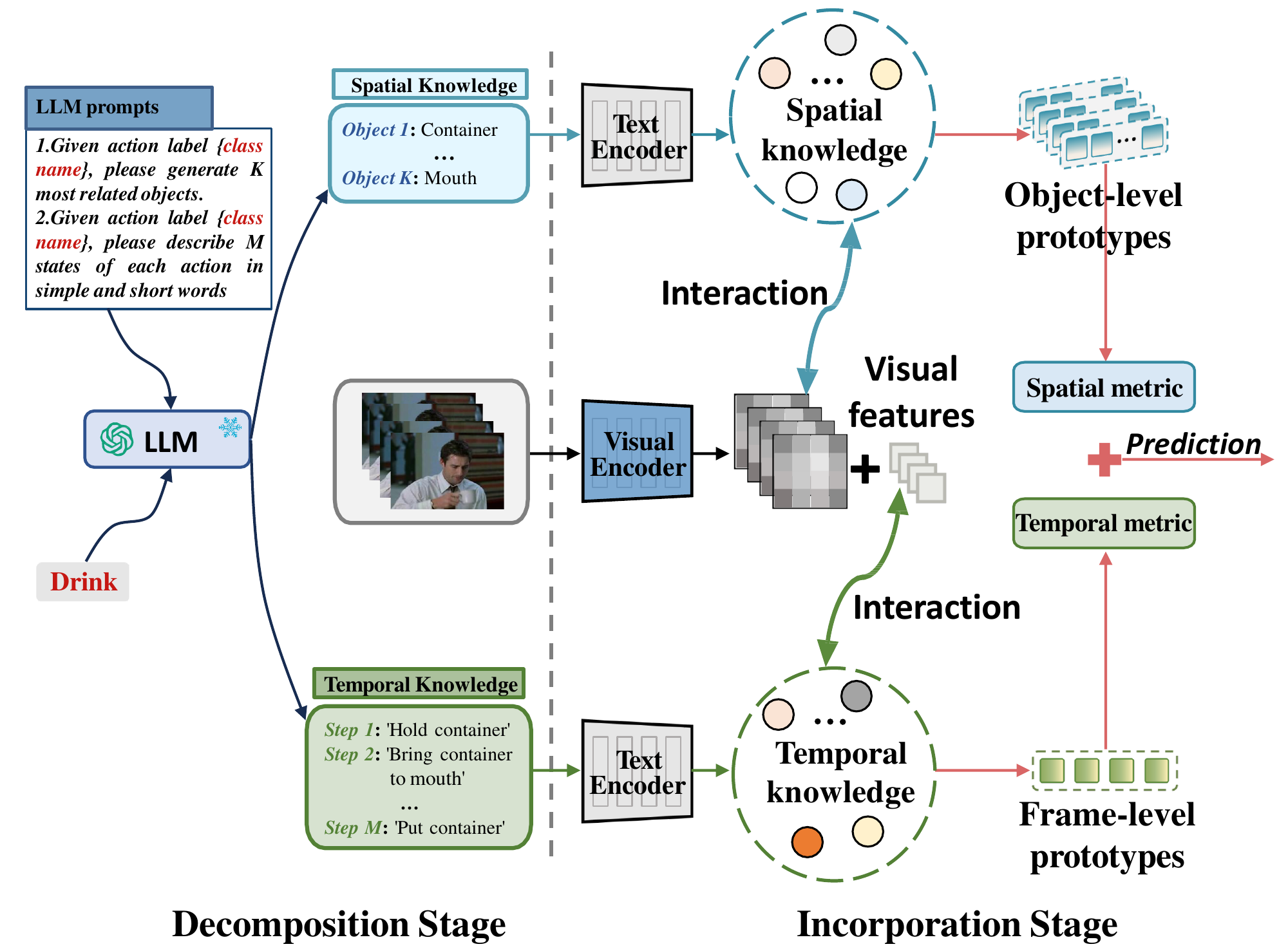} 
   \caption{\textbf{Our main idea}. \textsc{DiST} decomposes category names into diverse spatio-temporal knowledge, and makes use of such decoupled knowledge to guide the learning of object-level and frame-level prototypes, respectively.}
    \label{motivation}
    \vspace{-11pt}
\end{figure}
    
To address this issue, we study how to effectively collect and leverage action-related commonsense knowledge provided by large language models (LLMs), compensating for visual information and thus enhancing few-shot learning capacity. The LLM serves as a knowledge base to provide action-related background commonsense descriptions from complementary spatial and temporal perspectives, which we refer to as decoupled ``\textit{spatio-temporal prior knowledge}". Compared to vanilla categories, spatio-temporal prior knowledge \textbf{i)} makes up for missing context information to achieve semantic completeness, and \textbf{ii)}  transforms unseen categories into known commonsense descriptions, easily interpreted by pre-trained language models.

In light of the above, we develop a novel decomposition-incorporation framework for FSAR: \textbf{\textsc{DiST}}, which firstly decomposes vanilla category names into diverse spatio-temporal attribute descriptions, and then incorporates decoupled commonsense knowledge and visual features to guide the learning of object-level and frame-level prototypes. As shown in Fig.~\ref{motivation}, for the \textbf{decomposition stage}, we make use of LLMs to generate action-related commonsense descriptions (\ie, external contextual information and different steps of action) from coarse category names. Such comprehensive descriptions complement semantic contexts from spatial and temporal perspectives. In the \textbf{incorporation stage}, we propose Spatial/Temporal Knowledge Compensators (SKC/TKC), which incorporate decoupled prior knowledge and visual features to form discriminative object-level (spatial)  and frame-level (temporal) prototypes, respectively. Specifically, SKC aggregates important patches into compact object prototypes by patch-level cross-attention within each frame, and further guides object-level prototype learning with the assistance of spatial prior knowledge. These object-level prototypes filter out video noise and focus on informative image patches, which refer to the most class-related ones and correspond to key entities.  Meanwhile, TKC captures the temporal relationships between frame-level prototypes through inter-frame interaction, and then enables these prototypes to aggregate essential semantic information from temporal prior knowledge. The learned different-level prototypes can capture fine-grained spatial details and dynamic temporal information, so as to yield more accurate action recognition results.

Overall, our contributions are summarized as follows: 
\begin{itemize}
\item We pioneer the early exploration in making use of action-related prior knowledge for FSAR, and to achieve this end, we construct commonsense descriptions derived from LLMs in a spatiotemporal-decoupled manner. 
\item We propose a novel decomposition-incorporation framework that decouples vanilla category names into diverse spatial and temporal prior knowledge, and then incorporates these prior knowledge and visual features to learn object-level and frame-level prototypes in a dual way. 
\item We design Spatial/Temporal Knowledge Compensators (SKC/TKC) that inject decoupled prior knowledge into different-level prototypes (object-level and frame-level) to capture fine-grained spatial details and dynamic temporal information. 
\end{itemize}

To the best of our knowledge, we make the pioneering effort to explore the application of diverse spatio-temporal prior knowledge in FSAR, aiming to effectively provide semantic contexts for the learning of different-level prototypes from multiple perspectives. \textit{Different from} simply combining rich LLM-generated descriptions (\ie, only one sentence about actions) and frame-level features in a single branch, \textsc{DiST} conducts customized feature interaction for different branches to incorporate i) patch-level features and spatial knowledge; and ii) frame-level features and temporal knowledge, respectively (see Table~\ref{tab::abs}c in \S\ref{sec44} for ablation study). Such a framework can discover fine-grained spatial patterns and dynamic temporal patterns, hence producing more accurate few-shot results.  To comprehensively evaluate \textsc{DiST}, we conduct experiments on five gold-standard datasets (\ie, HMDB51~\cite{kuehne2011hmdb}, UCF101~\cite{soomro2012ucf101}, kinetics100~\cite{carreira2017quo}, SSv2-full~\cite{goyal2017something}, and SSv2-small~\cite{goyal2017something}). We empirically prove that \textsc{DiST} surpasses all existing state-of-the-art methods and yields solid performance gains (\textbf{1.7\%}-\textbf{6.8\%} accuracy) under the $5$-way $1$-shot setting.

\section{Related Work}\label{sec:related}
\subsection{Few-shot Image Classification} 
 The objective of few-shot image classification~\cite {fei2006one,guo2020broader} is to recognize new categories with a small number of annotated samples. Existing methods can be roughly categorized into three groups: \textbf{i)} \textit{Augmentation-based} methods~\cite{chen2019image,li2020adversarial,ye2021learning} exploit various augmentation strategies to alleviate the data scarcity dilemma, mainly including  spatial deformation~\cite{ratner2017learning} and  feature augmentation~\cite{chen2018semantic,chen2019multi}; \textbf{ii)} \textit{Optimization-based} methods~\cite{finn2017model,jamal2019task,rajeswaran2019meta,rusu2018meta} learn optimization states, like model initialization~\cite{finn2017model,jamal2019task}  or step sizes~\cite{rajeswaran2019meta,rusu2018meta}, to  update models with a few gradient steps; and \textbf{iii)} \textit{Metric-based} methods~\cite{yoon2019tapnet,snell2017prototypical,vinyals2016matching,ye2020few,sung2018learning,li2019finding}  learn a class representation (prototype) by averaging embeddings belonging to the same class, and classify query instances by measuring distances to these prototypes with respect to predefined~\cite{yoon2019tapnet,snell2017prototypical,vinyals2016matching,ye2020few} or learned~\cite{sung2018learning,li2019finding} distance metric.

Our work is more closely related to the metric-based methods~\cite{snell2017prototypical,vinyals2016matching}, whereas we focus on few-shot action recognition -- a more challenging task that requires handling videos encompassing a wealth of temporal information due to common and distinct patterns in nearby frames~\cite{zhao2022alignment}. With respect to this, we simultaneously consider local spatial details and dynamic temporal relations in videos by decomposing the matching process into complementary object-level prototype and frame-level prototype alignment.

\subsection{Few-shot Action Recognition} 
Few-shot Action Recognition (FSAR) is a challenging task with the goal of recognizing previously unseen action classes (\ie, query class) with a few labeled videos. Existing FSAR methods~\cite{zhu2018compound,wang2022hybrid,wang2023clip,boosting,thatipelli2022spatio} mainly belong to the metric-based meta-learning paradigm~\cite{snell2017prototypical}, which learns class (prototype) representation and performs prototype-query matching based on the learned distance metrics. These methods are mainly devoted to feature representation learning~\cite{thatipelli2022spatio,boosting} and matching strategy exploration~\cite{zhu2018compound,zhu2020label,wang2021semantic,cao2020few,perrett2021temporal,wang2022hybrid,boosting}. As a primary step, \textit{feature representation learning} helps models to learn expressive spatio-temporal features for further matching process. Recent approaches~\cite {thatipelli2022spatio,boosting} model temporal features through temporal attention operations~\cite{thatipelli2022spatio} or more fine-grained temporal-patch and temporal-channel interaction~\cite{boosting}, and further exploit low-level spatial features by patch-level information interaction within each frame or across frames~\cite{xing2023revisiting}. Some others make use of video features in a whole task (\ie, episode) to extract relevant discriminative patterns~\cite{boosting,wang2022hybrid} by a graph neural network~\cite{boosting} or attention-based relation modeling~\cite{wang2022hybrid}. For \textit{matching strategy exploration}, early works~\cite{zhu2018compound,zhu2020label,wang2021semantic} aggregate the frame features into a single video representation for video-level feature matching. Though straightforward, these methods suffer from suboptimal performance due to neglecting the temporal cues in videos. To address this limitation, the following approaches~\cite{cao2020few,perrett2021temporal,wang2022hybrid,boosting,wu2022motion} devise various temporal alignment metrics for frame-level feature matching. The temporal alignment methods explore different alignment granularities, \ie, frame-level alignment~\cite{cao2020few,wang2022hybrid}, segment-level alignment~\cite{perrett2021temporal}, and even frame-to-segment alignment~\cite{wu2022motion} that is also common in realistic video matching.

Recent works~\cite{wang2023clip,xing2023multimodal,wang2023few} have explored transferring knowledge from pre-trained vision-language models (\eg, CLIP~\cite{radford2021learning}) to improve the performance of FSAR models. Though promising, they heavily rely on semantically coarse or ambiguous category names as the semantic source to provide action-related context. Such context fails to provide sufficient background knowledge for comprehensive video understanding. In contrast, our \textsc{DiST} represents the first effort in FSAR to explicitly decouple class names into diverse spatio-temporal attribute descriptions (\ie, action-relevant knowledge) to complement semantic contexts. More significantly, such acquired decoupled knowledge is further injected into visual features to guide the learning of object-level and frame-level prototypes in a two-stream manner, so as to enable more accurate action recognition.
  
\subsection{Few-shot Learning with Semantic Information}  
Recent works on few-shot learning~\cite{xing2019adaptive,peng2019few,xu2022generating,chen2023semantic} integrate semantic information (provided by class labels) and visual information (extracted from visual observations) to represent a novel class. Based on the levels at which modality information fusion  occurs, these methods can be roughly categorized into three groups: \textbf{i)} \textit{Prototype-level} methods~\cite{xing2019adaptive,yan2022inferring}  model class (prototype) representation as a combination of visual prototypes and semantic prototypes obtained through word embeddings of class labels by attention mechanism~\cite{yan2022inferring} or adaptive fusion mechanism~\cite{xing2019adaptive}; \textbf{ii)} \textit{Classifier-level} methods~\cite{peng2019few} enable classifiers to predict novel categories by incorporating auxiliary  semantic information acquired from a graph convolutional network~\cite{wu2019simplifying}; and \textbf{iii)} \textit{Extractor-level} methods~\cite{zhang2023simple,chen2023semantic} treat semantic information as prompts to guide the adaptation of feature extractors, thereby encouraging the feature extractor to better focus on class-specific features.

Though impressive, they~\cite{xing2019adaptive,zhang2023simple,chen2023semantic} typically directly apply semantically coarse category names as auxiliary information at different levels to compensate for visual features, lacking high-quality background knowledge to discover novel visual concepts, therefore struggling with adapting to unseen categories. Our contribution is orthogonal to previous studies, as we advance FSAR regime in the aspect of collecting and leveraging high-quality spatio-temporal prior knowledge for FSAR. The LLM serves as an external knowledge source to provide action-related background commonsense descriptions (\ie, contextual background information and different procedural steps of actions). Then our work makes smart use of such prior knowledge to reduce redundant visual features and enhance the semantic distinction of different class prototypes.

\section{Method}
\subsection{Problem Formulation}
The goal of FSAR is to classify unlabeled test videos from novel classes given a few (\eg, one or five) samples per class. Under the few-shot setting, 
the model learns discriminative feature representation on training classes $\mathcal{D}_{base}$ and is evaluated on testing classes $\mathcal{D}_{novel}$ to emphasize its generalization ability on novel categories, where $\mathcal{D}_{base}\cap\mathcal{D}_{novel}=\emptyset$. In the training stage, following~\cite {tang2023m3net,wang2022hybrid,boosting}, we train a few-shot learning model in an episodic way. Here, each episode (\ie, a standard $M$-way $K$-shot episode task) is formed by sampling $M$ categories from $\mathcal{D}_{base}$. The $M$-way $K$-shot task consists of  $K$ labeled videos per class as the support set $\mathcal{S}$, and a fraction of the remaining samples from $M$ classes as the query set $\mathcal{Q}$. The episodic training for FSAR is achieved by minimizing, for each episode, the loss of the prediction on samples in the query set, given the support set. In the inference stage, we randomly sample episode tasks from $\mathcal{D}_{novel}$ for evaluation, and report average results over multiple episode tasks. 

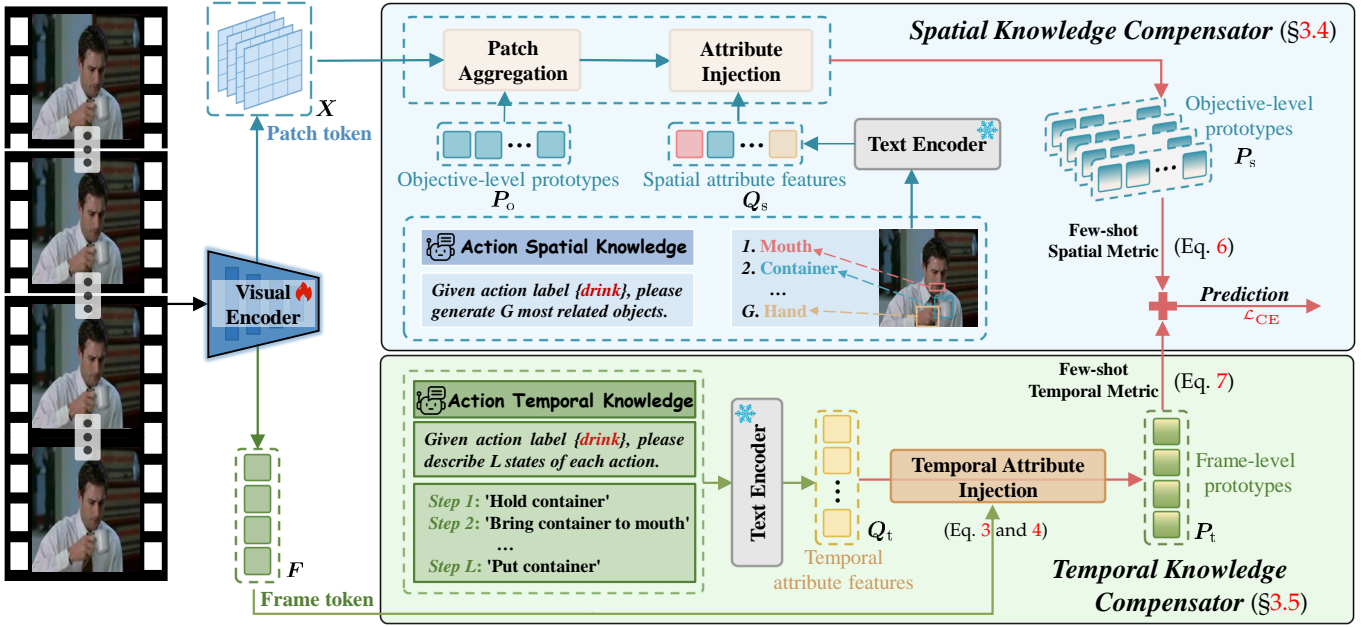
\begin{figure*}[!t]
    \centering
    \includegraphics[width=0.99\textwidth]{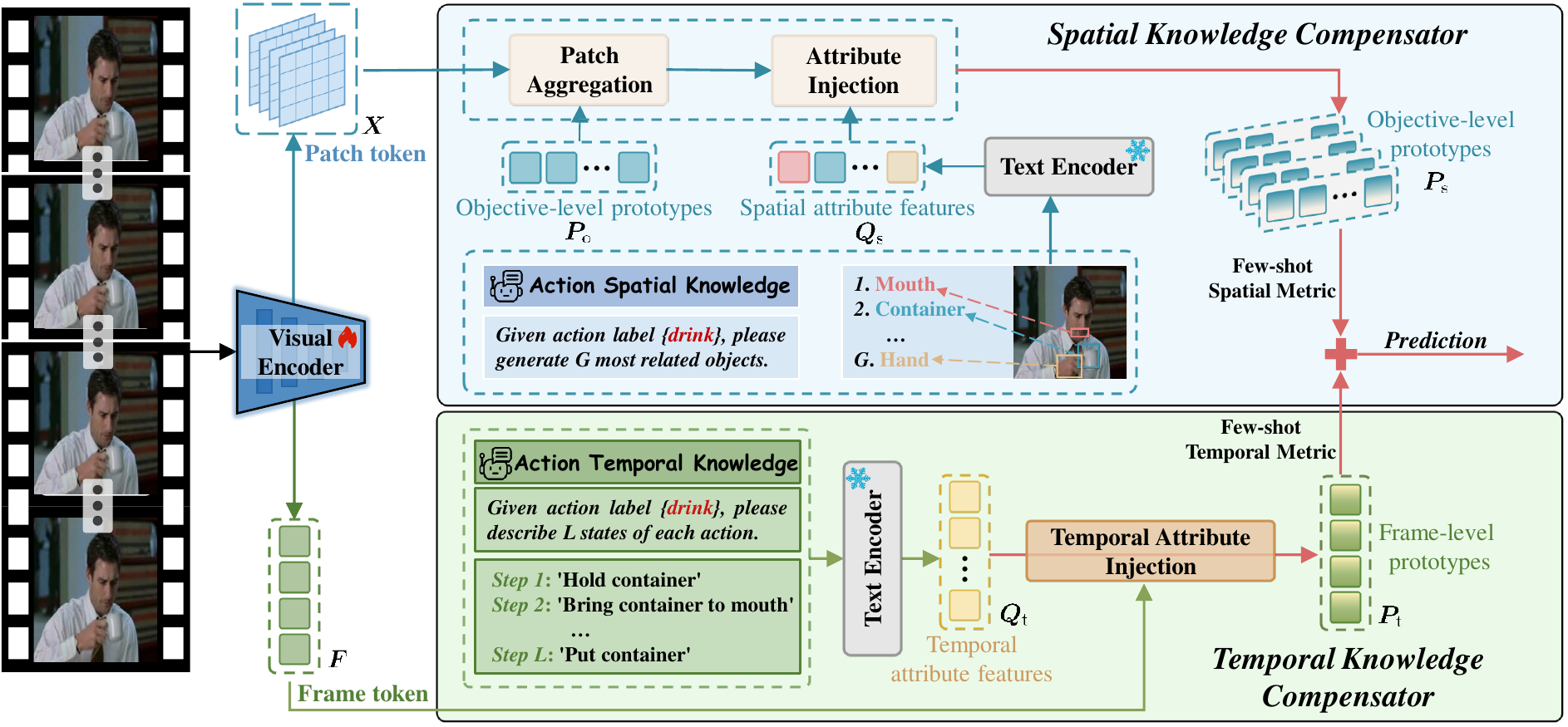} 
    \put(-30, 223.5){\small(\S\ref{SKC})}
    \put(-41, 7){\small(\S\ref{TKC})}
   \put(-70, 91){\footnotesize(Eq.~\ref{eq::temporal_metric})}
     \put(-70, 142){\footnotesize(Eq.~\ref{eq::spatial_metric})}
     \put(-157, 35){\scriptsize(Eq.~\ref{eq::TKC1} and \ref{gongshi})}
     \put(-44, 117){\tiny{\color{red}$\mathcal{L}_{\mathrm{CE}}$}}
    \caption{\textbf{Overview of \textsc{DiST}}. Video inputs are first processed by the visual encoder of CLIP to obtain initial patch-level and frame-level features. Then, we leverage LLM to decompose vanilla action names into action-related background knowledge (\S\ref{generation}). Furthermore, SKC/TKC incorporate decoupled prior knowledge and visual features to form discriminative object- and frame-level prototypes for spatial/temporal matching (\S\ref{SKC} and \S\ref{TKC}). Finally, we can combine spatial and temporal matching results to obtain the merged query prediction (\S\ref{metric}).}
    \vspace{-11pt}
    \label{overview}
\end{figure*}
\subsection{Overall Framework}
We introduce \textsc{DiST}, which collects and leverages action-related commonsense knowledge provided by LLMs to guide the learning of object- and frame-level prototypes for FSAR, as illustrated in Fig.~\ref{overview}. \textsc{DiST} consists of visual/text encoders and two knowledge compensators. The model takes the RGB frame sequence of length $T$ and corresponding action names as input. We first utilize the visual encoder of CLIP~\cite{radford2021learning} to get frame-level (\ie, class token in each frame) and patch-level feature representation,~\ie, $\bm{F} \in {\mathbb{R}}^{T \times C} $ and $\bm{X} \in {\mathbb{R}}^{T \times P \times C}$, where $P$ is the number of tokens in each frame. Besides, to provide external action-related prior knowledge, we prompt LLM with corresponding action names to generate decoupled spatial and temporal commonsense descriptions (\S\ref{generation}). These descriptions are fed into frozen text encoder of CLIP to obtain spatial and temporal attribute features,~\ie, $\bm{Q}_{\mathrm{s}} \in {\mathbb{R}}^{G \times C}$ and $\bm{Q}_{\mathrm{t}} \in {\mathbb{R}}^{L \times C}$. We further design two complementary modules to make use of decoupled spatio-temporal attributes: 1) Spatial Knowledge Compensator (SKC) (\S\ref{SKC}) injects spatial attributes into patch-level features to explicitly learn compact object-level prototypes for object-level prototype matching; 2) Temporal Knowledge Compensator (TKC) (\S\ref{TKC}) incorporates temporal attributes and frame-level features to inform frame-level prototypes for frame-level prototype matching. Finally, we can combine spatial and temporal feature matching scores to obtain the merged query prediction.

\subsection{Decoupled Spatio-temporal Attribute Generation}
\label{generation}
\noindent\textbf{Spatial Attribute Generation.} Naive category names provided limited commonsense knowledge to focus on action-related spatial contexts. Thus, we make use of prior knowledge in LLMs~\cite{brown2020language,ChatGPT} to generate detailed and informative spatial attributes for each category,~\ie, action-related object instances and environment. Specifically, taking the action category ``\textit{drink}" as an example, to obtain spatial attribute descriptions, $\!$we prompt ChatGPT~\cite{ChatGPT} by ``\textit{Given action label \{drink\}, please generate \{$G$}\textit{\} most related objects for each class.}", where $G$ is empirically set to $6$ (ablation study in Table~\ref{tab:promptnum}). This prompt returns a set $\mathcal{A}$ with $G$ spatial attribute descriptions, such as ``\textit{container; mouth; hand; ...}". Then we encode these spatial attributes via frozen CLIP text encoder to get spatial attribute features $\bm{Q}_{\mathrm{s}}\!\in\!{\mathbb{R}}^{G \times C}$.  

\noindent\textbf{Temporal Attribute Generation}. Prior researches~\cite{wang2023clip,xing2023multimodal} apply semantically coarse category names as auxiliary information to guide temporal feature learning. However, such context provided by action names is too limited to provide enough temporal context for action recognition. Thus, we propose to utilize the abundant prior knowledge in LLMs~\cite{brown2020language,ChatGPT} to expand the coarse action names. Temporal attribute descriptions generated by LLM are a collection of multiple atomic actions, which describe the temporal evolution of an action. Concretely, taking the action category ``\textit{drink}" as an example, to obtain temporal attribute descriptions, we prompt ChatGPT~\cite{ChatGPT} by ``\textit{Given action label \{drink\}, please describe \{}$L$\textit{\} states of each action in simple and short words.}", where we empirically set  $L$ to $3$ (Table~\ref{tab:promptnum}). This prompt always returns a set $\mathcal{B}$ with $L$ temporal attribute descriptions, such as ``\textit{Hold container; Bring container to mouth; Put container; ...}", which decompose one action class into multiple atomic actions in a step-by-step manner. Then we adopt the off-the-shelf text encoder of CLIP to encode these descriptions and obtain temporal attribute features $\bm{Q}_{\mathrm{t}} \in {\mathbb{R}}^{L\times C}$.

\subsection{Spatial Knowledge Compensator}
\label{SKC}
\begin{figure*}[!t]
    \centering
   \includegraphics[width=0.98\textwidth]{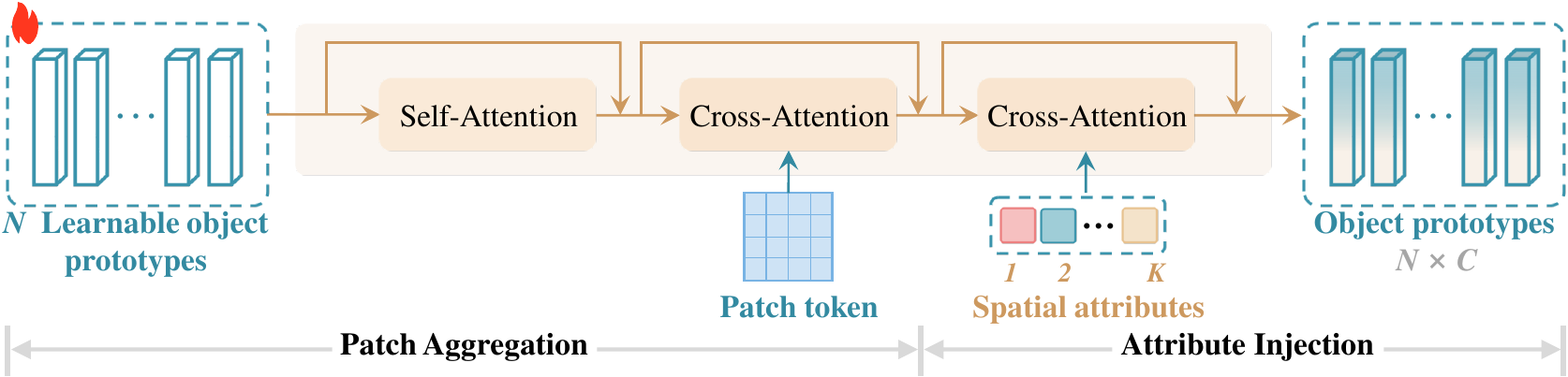}
    \caption{Illustration of Spatial
Knowledge Compensator (SKC) (\S\ref{SKC}). SKC aims to learn discriminative object-level prototypes in a sparse aggregation manner via patch aggregation and attribute injection. }
    \label{SPI}
\end{figure*}

Previous methods exploit spatial features by patch-level information interaction within each frame or across frames~\cite{thatipelli2022spatio,xing2023revisiting}. However, this leads to two issues: 1) Too many irrelevant patch tokens bring redundant information, interfering with further spatial feature matching; 2) They fail to focus on important objects without the guidance of spatial prior knowledge. Therefore, we investigate how to better incorporate spatial attributes (\S\ref{generation}) and patch-level visual features into compact object-level prototypes to highlight potential target objects. To this end, as showcased in Fig.~\ref{SPI}, proposed Spatial Knowledge Compensator (SKC)  summarizes discriminative spatial patterns via aggregating patch-level features into compact object-level prototypes (\ie, patch aggregation), and further delivers the union of spatial attributes and such object-level prototypes to focus on semantically relevant object regions guided by commonsense knowledge (\ie, attribute injection).

\noindent\textbf{Patch Aggregation.} We first introduce a set of learnable object-level prototypes to aggregate image content and highlight potential target objects. The prototypes are randomly initialized embeddings and represented as $\bm{P}_\mathrm{o} \in {\mathbb{R}}^{N \times C}$, where $N$ is the number of object prototypes.  Firstly, a self-attention layer is adopted for the $N$ object prototypes to interact with each other in each frame. Then, these prototypes aim to adaptively aggregate action-related or object-related key patches in a sparse manner by patch-level cross-attention within each frame. Specifically, for patch tokens $\bm{X}^{l} \in {\mathbb{R}}^{P \times C}$ in $l$-th frame, the process can be defined as:
\vspace{-3pt}
\begin{equation}
\label{eq::SKC1}
         \hat{\bm{P}_{\mathrm{o}}} = \mathrm{Softmax}(\bm{P}_{\mathrm{o}}\bm{K}_{\mathrm{p}}^{\top})\bm{V}_{\mathrm{p}} + \bm{P}_{\mathrm{o}},
\end{equation}
where $\bm{K}_{\mathrm{p}}$ and $\bm{V}_{\mathrm{p}}$ are the linear transformation features of patch tokens $\bm{X}^{l}$. This allows the object-level prototypes to capture discriminative spatial patterns. 

 \noindent\textbf{Attribute Injection.} To further encourage object prototypes to focus on action-related spatial context information, we deliver the union of spatial attribute knowledge and such object-level prototypes to discover fine-grained spatial patterns via  the attention mechanism as follows:
\vspace{-3pt}
\begin{equation}
\setlength\belowdisplayskip{2pt}
\label{eq::SKC2}
         \bm{P}_{\mathrm{s}} = \mathrm{Softmax}(\hat{\bm{P}_{\mathrm{o}}}\bm{K}_{\mathrm{q}}^{\top})\bm{V}_{\mathrm{q}} + \hat{\bm{P}_{\mathrm{o}}},
\end{equation}
where $\bm{K}_{\mathrm{q}}$ and $\bm{V}_{\mathrm{q}}$ are linear transformation features of spatial attribute features $\bm{Q}_{\mathrm{s}}$, $\bm{P}_{\mathrm{s}} \in {\mathbb{R}}^{ N \times C}$ is learned diverse object prototypes, where $N$ is the number of prototypes. Note that the operation of object prototypes in each frame is the same. By exchanging information for visual features and attribute features respectively, learned prototypes filter out redundant information in videos and capture spatial details.

\begin{algorithm}[t]
    \caption{Pseudo-code of object-level and frame-level prototype matching in a PyTorch-like style.}
    \label{alg:code}
         \definecolor{codeblue}{rgb}{0.25,0.5,0.5}
    \definecolor{codered}{rgb}{0.80,0.25,0.25}
    \lstset{
      basicstyle=\fontsize{7.6pt}{7.6pt}\ttfamily\bfseries,
      commentstyle=\fontsize{7.6pt}{7.6pt}\color{codeblue},
      keywordstyle=\fontsize{7.6pt}{7.6pt}\color{codered},
    }
\begin{lstlisting}[language=python]
# q_obj_p: query object prototype  (B_q x T x N x C) 
# s_obj_p: support object prototype  (B_s x T x N x C)  
# q_frm_p: query frame prototype (B_q x T x C)
# s_frm_p: support frame prototype (B_s x T x C)
# B_q: number of query videos
# B_s: number of support videos
# T: number of frames
# N: number of object prototypes in each frame
# OTAM: temporal metric
   
def matching(q_obj_p, s_obj_p, q_frm_p, s_frm_p):
    # Object-level prototype matching
    q_obj_p = q_obj_p.reshape(B_q x T x N, C)  
    s_obj_p = s_obj_p.reshape(B_s x T x N, C)  
    obj_sim = cos_sim(q_obj_p, s_obj_p) 
    obj_dist = 1 - obj_sim 
    #  obj_dist: (B_q x T) x (B_s x T) x N x N
    obj_dist = rearrange(obj_dist) 
    obj_fdist = obj_dist.min(3)[0].sum(2) \ 
                + obj_dist.min(2)[0].sum(2)  
    obj_fdist = rearrange(obj_fdist) # B_q x B_s x T x T
    obj_logits = obj_fdist.min(3)[0].mean(2) + \
                 obj_fdist.min(2)[0].mean(2) # B_q x B_s 
   
    # Frame-level prototype matching
    q_frm_p = q_frm_p.reshape(B_q x T, C)  
    s_frm_p = s_obj_p.reshape(B_s x T, C) 
    frm_sim = cos_sim(q_frm_p, s_frm_p) 
    frm_dist = 1 - frm_sim 
    frm_dist = rearrange(frm_dist)  # B_q x B_s x T x T
    frm_logits = OTAM(frm_dist) # B_q x B_s
        
    return obj_logits, frm_logits
\end{lstlisting}
\end{algorithm}
   
\subsection{Temporal Knowledge Compensator}
\label{TKC}
How to incorporate temporal attribute features and frame-level visual features is essential for better FSAR performance since temporal prior knowledge can enable the model to understand dynamic semantics in actions. Thus, our Temporal Knowledge Compensator (TKC) aggregates essential semantic information by injecting temporal prior knowledge into visual features, facilitating a comprehensive understanding of spatio-temporal dynamics.

Specifically, we obtain global semantic vector $\bm{p}_{\mathrm{g}} \in {\mathbb{R}}^{1 \times C}$ by pooling temporal attribute features, and add it to frame-level features $[\bm{f}_{1},\bm{f}_{2},...,\bm{f}_{T}] \in {\mathbb{R}}^{T \times C}$:
\vspace{-3pt}
\begin{equation}
\setlength\belowdisplayskip{2pt}
\label{eq::TKC1}
          \bm{F}_{\mathrm{q}} = [\bm{f}_{1}+\bm{p}_{\mathrm{g}},...,\bm{f}_{T}+\bm{p}_{\mathrm{g}}],
\end{equation}
where  $\bm{F}_{\mathrm{q}} \in {\mathbb{R}}^{T \times C}$ is the obtained frame-level prototypes,  which incorporate overall semantic information. The frame-level prototypes further aggregate temporal context information from temporal prior knowledge via vision and attribute cross-attention mechanism. Then the frame prototypes are fed into the temporal transformer~\cite{wang2022hybrid} to capture the temporal relationships between frame-level prototypes. This is given by
\vspace{-3pt}
\begin{equation}
\setlength\belowdisplayskip{2pt}
         \bm{P}_{\mathrm{t}} = \mathtt{Tformer}(\mathrm{Softmax}(\bm{F}_{\mathrm{q}}\bm{K}_{\mathrm{t}}^{\top})\bm{V}_{\mathrm{t}} + \bm{F}_{\mathrm{q}}),
         \label{gongshi}
\end{equation}
where $\bm{K}_{\mathrm{t}}$ and $\bm{V}_{\mathrm{t}}$ are the linear transformation features of temporal attribute features $\bm{Q}_{\mathrm{t}}$, $\mathtt{Tformer}$ is the temporal transformer~\cite{wang2022hybrid}, and $\bm{P}_{\mathrm{t}} \in {\mathbb{R}}^{T \times C} $ is the frame-level prototypes capturing action dynamic information. In this way, the learned frame-level prototypes can adaptively perceive temporal changes and encode the action temporal context with the guidance of temporal knowledge.
\begin{table*}[!t]
\caption{\textbf{Quantitative comparison results on HMDB51~\cite{kuehne2011hmdb} and UCF101~\cite{soomro2012ucf101}} (see \S\ref{sec43}). The experiments are conducted under the $5$-way $K$-shot. ``INet-RN50" denotes ResNet-50 pre-trained on ImageNet. We highlight \textbf{best}, \underline{second best} results.}
\label{table1}
\centering
\small
\resizebox{0.98\textwidth}{!}{
\setlength\tabcolsep{8pt}
\renewcommand\arraystretch{1.2}
\begin{tabular}
{rl||c|ccc|ccc}
\thickhline
\rowcolor{mygray}   &&

&\multicolumn{3}{c|}{{HMDB51}} 
&\multicolumn{3}{c}{{UCF101}}\\ 
  \rowcolor{mygray}
    \multicolumn{2}{c||}{\multirow{-2}{*}{Method}} & \multicolumn{1}{c|}{\multirow{-2}{*}{\hspace{0.5mm} Pre-training \hspace{0.5mm}}}  & $1$-shot (\%)  & $3$-shot (\%) & $5$-shot (\%) & $1$-shot (\%) & $3$-shot (\%)  & $5$-shot (\%) \\
\hline
 \hline
 ProtoNet~\cite{snell2017prototypical}\!\!\!\!\!\!&\!\!\pub{NeurIPS17}              & INet-RN50   & $54.2$  & -  & $68.4$    & $74.0$  & -    & $89.6$     \\
ARN~\cite{zhang2020few}\!\!\!\!\!\!&\!\!\pub{ECCV20}              & C3D  & $45.5$  & -  & $60.6$    & $66.3$  & -    & $83.1$     \\
OTAM~\cite{cao2020few}\!\!\!\!\!\!&\!\!\pub{CVPR20}              & INet-RN50  & $54.5$  & $65.7$  & -     & $79.9$  & $87.0$    & -     \\
AmeFu-Net~\cite{fu2020depth}\!\!\!\!\!\!&\!\!\pub{MM20}       & INet-RN50     & $60.2$    & $-$      & $75.5$   & $85.1$      & $-$   & $95.5$  \\  
Lite-MKD~\cite{liu2023lite}\!\!\!\!\!\!&\!\!\pub{MM23}       & INet-RN50    & $66.9$    & $-$      & $74.7$  & $85.3$      & $-$   & $96.8$   \\
TRX~\cite{perrett2021temporal}\!\!\!\!\!\!&\!\!\pub{CVPR21}              & INet-RN50  & $53.1$  & $66.8$  & $75.6$  & $78.2$  & $92.4$  & $96.1$  \\
TA$^{2}$N~\cite{li2022ta2n}\!\!\!\!\!\!&\!\!\pub{AAAI22}       &  INet-RN50     & $59.7$      & -       & $73.9$ & $81.9$    & -       & $95.9$    \\
MTFAN~\cite{wu2022motion}\!\!\!\!\!\!&\!\!\pub{CVPR22}             & INet-RN50  & $59.0$    & -     & $74.6$  & $84.8$  & -     & $95.1$  \\
HyRSM~\cite{wang2022hybrid}\!\!\!\!\!\!&\!\!\pub{CVPR22}             & INet-RN50  & $60.3$  & $71.7$  & $76.0$    & $83.9$  & $93.0$    & $94.7$  \\
STRM~\cite{thatipelli2022spatio}\!\!\!\!\!\!&\!\!\pub{CVPR22}              & INet-RN50  & $52.3$  & $67.4$  & $77.3$  & $80.5$  & $92.7$  & $96.9$  \\
CPM~\cite{huang2022compound}\!\!\!\!\!\!&\!\!\pub{ECCV22}              & INet-RN50  & $60.1$  & -     & -     & $71.4$  & -     & -     \\
SloshNet~\cite{xing2023revisiting}\!\!\!\!\!\!&\!\!\pub{AAAI23}              & INet-RN50  & $-$  & $-$  & $77.5$  & $-$  & $-$    & $97.1$  \\
HCL~\cite{zheng2022few}\!\!\!\!\!\!&\!\!\pub{ECCV22}              & INet-RN50  & $59.1$  & $71.2$  & $76.3$  & $82.6$  & $91.0$    & $94.5$  \\
MoLo~\cite{wang2023molo}\!\!\!\!\!\!&\!\!\pub{CVPR23}              & INet-RN50  & $60.8$  & $72.0$    & $77.4$  & $86.0$    & $93.5$  & $95.5$  \\
GgHM~\cite{boosting}\!\!\!\!\!\!&\!\!\pub{ICCV23}             & INet-RN50  & $61.2$  & -    & $76.9$  & $85.2$    & -  & $96.3$  \\
CLIP-FSAR~\cite{wang2023clip}\!\!\!\!\!\!&\!\!\pub{IJCV24}         & CLIP-RN50  & $69.2$  & $77.6 $ & $80.3$  & $91.3$  & $95.1$  & $97.0$    \\
CapFSAR~\cite{wang2023few}\!\!\!\!\!\!&\!\!\pub{Arxiv23}       & BLIP-ViT-B  & $65.2$  & -     & $78.6$  & $93.3$  & -     & $97.8$  \\
MVP-shot~\cite{qu2024mvp}\!\!\!\!\!\!&\!\!\pub{Arxiv24}       & CLIP-ViT-B  & $69.2$ &-&$80.3$  & $91.3$  & -     & $97.0$  \\
CLIP-Freeze~\cite{radford2021learning}\!\!\!\!\!\!&\!\!\pub{ICML21}       & CLIP-ViT-B  & $58.2$  & $72.7$     & $77.0$  & $89.7$  & $94.3$     & $95.7$  \\
TEAM~\cite{lee2025temporal}\!\!\!\!\!\!&\!\!\pub{CVPR25}             & INet-ViT-B  & $70.9$  & -    & $85.5$  & $94.5$    & -  & $98.8$  \\
CLIP-FSAR~\cite{wang2023clip}\!\!\!\!\!\!&\!\!\pub{IJCV24}         & CLIP-ViT-B & \underline{$75.8$}  & \underline{$84.1$}  & \underline{$87.7$}  & \underline{$96.6$}  & \underline{$98.4$}  & \underline{$99.0$}    \\
\hline \hline
\multicolumn{2}{c||}{\textsc{DiST}  \textbf{(Ours)}}                  & CLIP-ViT-B & $\mathbf{82.6}$\tiny{$\pm$0.3}  & $\mathbf{87.1}$\tiny{$\pm$0.3}  & $\mathbf{88.7}$\tiny{$\pm$0.1}  & $\mathbf{98.3}$\tiny{$\pm$0.2}  & $\mathbf{99.0}$\tiny{$\pm$0.2}    & $\mathbf{99.2}$\tiny{$\pm$0.1}  \\
\hline
\end{tabular}
  }
\end{table*}

\subsection{Few-shot Metric}
\label{metric}
\noindent \textbf{Few-shot Spatial Metric.} To conduct spatial feature matching between videos, we propose an object-level prototype matching strategy based on the bidirectional Hausdorff Distance~\cite{wang2022hybrid}, which calculates the distances between query object-level prototypes and support object-level prototypes from the set matching perspective. Specifically, given query object-level prototypes $\bm{P}_{\mathrm{s}} \in {\mathbb{R}}^{T \times N \times C}$  and support object-level prototypes $\hat{\bm{P}_{\mathrm{s}}} \in {\mathbb{R}}^{T \times N \times C}$, we apply a bidirectional Mean Hausdorff
metric  to obtain a frame-level distance matrix $\bm{\hat{D}}={[d_{ij}]}_{T \times T} \in {\mathbb{R}}^{T \times T}$ as:
\vspace{-3pt}
\begin{equation}
\setlength\belowdisplayskip{2pt}
 \begin{split}
    d_{ij} &= \frac{1}{N} \sum_{{\bm{p}_{i,k}^s} \in {\bm{p}_{i}^s}}(\min_{\hat{\bm{p}}_{j,l}^s \in \hat{\bm{p}}_{j}^s} \begin{Vmatrix}
       {\bm{p}_{i,k}^s}- \hat{\bm{p}}_{j,l}^s
   \end{Vmatrix}) \\ 
   &+ \frac{1}{N} \sum_{\hat{\bm{p}}_{j,l}^s \in \hat{\bm{p}}_{j}^s}(\min_{{\bm{p}_{i,k}^s} \in {\bm{p}_{i}^s}} \begin{Vmatrix}
       \hat{\bm{p}}_{j,l}^s- {\bm{p}_{i,k}^s}
   \end{Vmatrix}),
 \end{split}
\end{equation}
where ${\bm{p}_{i,j}^s}$ and $\hat{\bm{p}}_{i,j}^s$ are the $j$-th query and support object-level prototypes in $i$-th frame, respectively. Then, for the frame-level distance matrix $\bm{\hat{D}} \in {\mathbb{R}}^{T \times T}$, we find the smallest distance across the frame sequences, which gives a more confident probability of spatial feature matching. Finally, the spatial metric  can be formulated as:
\vspace{-3pt}
\begin{equation} \label{eq::spatial_metric} \begin{split} {\mathcal{D}}_{\mathrm{s}} = \underbrace{\frac{1}{T} \sum\nolimits_{i=1}^{T} \left( \min_{j \in \{1...T\}} d_{ij} \right)}_{\text{Query-to-Support}} + \underbrace{\frac{1}{T} \sum\nolimits_{j=1}^{T} \left( \min_{i \in\{1...T\}} d_{ij} \right)}_{\text{Support-to-Query}}, \end{split} \end{equation}
where the indices $i$ and $j$ refer to query and support frame indices respectively, and $d_{ij}$ represents the frame-level distance distance between the $i$-th query frame and the $j$-th support frame. The first term calculates the average of the minimum distances from each frame $i$ in the query video to the most semantically similar frame in the support video. Inversely, the second term calculates the average of the minimum distances from each frame $j$ in the support video to the most similar frame in the query video.

\noindent \textbf{Few-shot Temporal Metric.}  After learning frame-level prototypes of support and query videos in a few-shot task, like in prior works~\cite{cao2020few,wang2023clip}, we obtain support-query matching results by applying the temporal alignment metric:
\vspace{-3pt}
\begin{equation}
\setlength\belowdisplayskip{2pt}
\label{eq::temporal_metric}
{\mathcal{D}}_{t} = \mathtt{Metric}(\bm{P}_{\mathrm{t}}, \hat{\bm{P}_{\mathrm{t}}}),
\end{equation}
where $\bm{P}_{\mathrm{t}} \in {\mathbb{R}}^{T \times C}$ represents  query frame-level prototypes, $\hat{\bm{P}_{\mathrm{t}}} \in {\mathbb{R}}^{T \times C}$ is  support frame-level prototypes,and
$\mathtt{Metric}$ denotes the  OTAM~\cite{cao2020few} metric by default. We formulate the distance between support and query videos as the weighted sum of the distances obtained by the few-shot spatial metric and few-shot temporal metric:
\vspace{-3pt}
\begin{equation}
\setlength\belowdisplayskip{2pt}
{\mathcal{D}} = {\mathcal{D}}_{\mathrm{t}} + {\alpha}{\mathcal{D}}_{\mathrm{s}},
\label{gongshi2}
\end{equation}
where ${\alpha}$ is a coefficient parameter. Our proposed spatial and temporal matching strategies combine the advantages of frame- and object-level prototype matching to cope with appearance-centric and motion-centric actions.

Algorithm~\ref{alg:code} presents the pseudo-code for spatial and temporal feature matching in a PyTorch-like style. Leveraging the acquired action-related prior knowledge, we decompose the video matching process into object-level prototype matching and frame-level prototype matching. This dual-level matching scheme enables more fine-grained alignment between query and support videos, boosting recognition performance under few-shot settings. 

 Following previous works~\cite{cao2020few,perrett2021temporal,wang2022hybrid}, we minimize cross-entropy loss $\mathcal{L}_{\mathrm{CE}}$ over the support-query distances based on the ground-truth labels to end-to-end train \textsc{DiST}. For few-shot inference, total support-query distance in Eq.~\ref{gongshi2} is employed as logits to produce final query prediction.

 \subsection{Discussion on LLM-generated universal attributes}
 Although \textsc{DiST} makes use of LLMs to generate spatio-temporal attributes, these attributes are universal and class-level, rather than instance-specific. This \textit{differs} from multimodal LLMs (MLLMs), which provide fine-grained and instance-specific captions. Thus, certain object predicted by LLMs (\eg, \textit{stool} in the action of drinking water) may not appear in every video instance. This raises concerns about the impact of irrelevant attributes on model performance.

To address this issue, we adopt \textit{cross-attention mechanisms (\ie, Spatial and Temporal Attribute Injection) that dynamically align LLM-generated attributes with visual features}. This allows \textsc{DiST} to attend selectively to relevant tokens while suppressing semantically unrelated ones. The attention weights are learned end-to-end, enabling robust grounding of class-level knowledge without requiring exact spatial and temporal alignment. Such model design not only \textit{mitigates the influence of noisy attributes}, but also \textit{promotes better generalization across varying video contexts} because our attributes provide universal, category-level knowledge about actions to capture semantic essence, free from visual biases.

Moreover, our LLM-generated universal attributes are \textit{highly efficient and reusable}: these attributes are generated once per action category, incurring minimal cost compared to inference-time MLLMs. In the future, we also consider incorporating object relevance estimation to further filter out potentially noisy concepts based on dataset-specific priors.

\begin{table*}[!t]
\caption{\textbf{$_{\!}$Quantitative results$_{\!}$ on$_{\!}$ Kinetics$_{\!}$~\cite{carreira2017quo}, SSv2~\cite{goyal2017something} and$_{\!}$ SSv2-small~\cite{goyal2017something}} (see$_{\!}$ \S\ref{sec43}). The$_{\!}$ experiments are$_{\!}$ conducted$_{\!}$ under$_{\!}$ the$_{\!}$ $5$-way$_{\!}$ $K$-shot. ``INet-RN50"$_{\!}$ denotes$_{\!}$ ResNet-50$_{\!}$ pre-trained$_{\!}$ on$_{\!}$ ImageNet. We highlight \textbf{best}, \underline{second best} results.  
}
\centering
\small
\resizebox{0.98\textwidth}{!}{
\setlength\tabcolsep{8pt}
\renewcommand\arraystretch{1.2}
\begin{tabular}
{rl||c|cc|cc|cc}
\thickhline
\rowcolor{mygray}
  && &\multicolumn{2}{c|}{{Kinetics}} &\multicolumn{2}{c|}{{SSv2}}
&\multicolumn{2}{c}{{SSv2-small}}\\
\rowcolor{mygray} 
\multicolumn{2}{c||}{\multirow{-2}{*}{Method}}  & \multicolumn{1}{c|}{\multirow{-2}{*}{Pre-training}} & $1$-shot (\%)   & $5$-shot (\%)& $1$-shot (\%)   & $5$-shot (\%) & $1$-shot (\%)   & $5$-shot (\%) \\
\hline \hline
ProtoNet~\cite{snell2017prototypical}\!\!\!\!\!\!&\!\!\pub{NeurIPS17}               & INet-RN50        & $64.5$       & $77.9$ &   -     & -  & -       & -     \\
ARN~\cite{zhang2020few}\!\!\!\!\!\!&\!\!\pub{ECCV20}               & C3D        & $63.7$       & $82.4$ &   -     & -  & -       & -     \\
OTAM~\cite{cao2020few}\!\!\!\!\!\!&\!\!\pub{CVPR20}              & INet-RN50  & $73.0$      & $85.8$&   $42.8$     & $52.3$  & $36.4$    & $48.0$    \\
Lite-KMD~\cite{liu2023lite}\!\!\!\!\!\!&\!\!\pub{MM23}      &  INet-RN50  & $75.0$    & $85.7$      & $55.7$       & $69.9$ &-& -  \\ 
TRX~\cite{perrett2021temporal}\!\!\!\!\!\!&\!\!\pub{CVPR21}                & INet-RN50  & $63.6$    & $85.9$&    $42.0$    & $64.6$  & $36.0$      & $56.7$ \\
TA$^{2}$N~\cite{li2022ta2n}\!\!\!\!\!\!&\!\!\pub{AAAI22}       &  INet-RN50   & $72.8$          & $85.8$               & -   & -      & -       & -   \\
MTFAN~\cite{wu2022motion}\!\!\!\!\!\!&\!\!\pub{CVPR22}             & INet-RN50  & $74.6$      & $87.4$&   $45.7$     &  $60.4$ & -          & -     \\
HyRSM~\cite{wang2022hybrid}\!\!\!\!\!\!&\!\!\pub{CVPR22}            & INet-RN50  & $73.7$   & $86.1$ &   $54.3$     & $69.0$ & $40.6$       & $56.1$  \\
STRM~\cite{thatipelli2022spatio}\!\!\!\!\!\!&\!\!\pub{CVPR22}            & INet-RN50  & $62.9$    & $86.7$ &   $43.1$     & $68.1$ & $37.1$    & $55.3$  \\
CPM~\cite{huang2022compound}\!\!\!\!\!\!&\!\!\pub{ECCV22}               & INet-RN50  & $73.3$       & - &   $49.3$     &  $66.7$   & -          & -     \\
HCL~\cite{zheng2022few}\!\!\!\!\!\!&\!\!\pub{ECCV22}               & INet-RN50  & $73.7$    & $85.8$ &    $47.3$    & $64.9$ & $38.9$    & $55.4$  \\
SloshNet~\cite{xing2023revisiting}\!\!\!\!\!\!&\!\!\pub{AAAI23}               & INet-RN50  & $-$    & $87.0$ &    $46.5$    & $68.3$ & $-$    & $-$  \\
MoLo~\cite{wang2023molo}\!\!\!\!\!\!&\!\!\pub{CVPR23}              & INet-RN50  & $74.0$      & $85.6$ &  $56.6$      & $70.6$ & $42.7$    & $56.4$  \\
CLIP-FSAR~\cite{wang2023clip}\!\!\!\!\!\!&\!\!\pub{IJCV24}         & CLIP-RN50  & $87.6$   & $91.9$ &   $58.1$     & $62.8$ & $52.0$        & $55.8$  \\
CapFSAR~\cite{wang2023few}\!\!\!\!\!\!&\!\!\pub{Arxiv23}      & BLIP-ViT-B  & $84.9$      & $93.1$ &    $51.9$    & $68.2$ & $45.9$       & $59.9$  \\
MVP-shot~\cite{qu2024mvp}\!\!\!\!\!\!&\!\!\pub{Arxiv24}      & CLIP-ViT-B  & $91.0$      & $95.1$ &    -    & - & $55.4$       & $62.0$  \\
CLIP-Freeze~\cite{radford2021learning}\!\!\!\!\!\!&\!\!\pub{ICML21}       & CLIP-ViT-B  & $78.9$      & $91.9$&   $30.0$     & $42.4$ & $29.5$       & $42.5$  \\
TEAM~\cite{lee2025temporal}\!\!\!\!\!\!&\!\!\pub{CVPR25}        & INet-ViT-B & $83.3$    & $92.9$ &    -    & -   & $47.2$    & \underline{$63.1$}  \\
CLIP-FSAR~\cite{wang2023clip}\!\!\!\!\!\!&\!\!\pub{IJCV24}        & CLIP-ViT-B & \underline{$89.7$}    & \underline{$95.0$} &    \underline{$61.9$}    & \underline{$72.1$}   & \underline{$54.5$}    & $61.8$  \\
\hline \hline
\multicolumn{2}{c||}{\textsc{DiST} \textbf{(Ours)}}                       & CLIP-ViT-B & $\mathbf{92.7}$\tiny{$\pm$0.3}    & $\mathbf{95.5}$\tiny{$\pm$0.1}&  $\mathbf{64.2}$\tiny{$\pm$0.2}      & $\mathbf{75.2}$\tiny{$\pm$0.2}  & $\mathbf{57.5}$\tiny{$\pm$0.3}   & $\mathbf{63.4}$\tiny{$\pm$0.1}  \\
\hline
\end{tabular}}
\label{table2}
\end{table*}

\begin{table*}[!t]
\caption{\textbf{Quantitative comparison results on UCF101~\cite{soomro2012ucf101}, HMDB51~\cite{kuehne2011hmdb}, Kinetics~\cite{carreira2017quo}, SSv2~\cite{goyal2017something} and SSv2-small~\cite{goyal2017something}} (see \S\ref{sec43}) by combining few-shot and zero-shot results. }
\centering
\small
\resizebox{0.98\textwidth}{!}{
\setlength\tabcolsep{6pt}
\renewcommand\arraystretch{0.95}
\begin{tabular}
{rl||cc|cc|cc|cc|cc}
\thickhline
\rowcolor{mygray} &&\multicolumn{2}{c|}{{HMDB51 (\%)}}&\multicolumn{2}{c|}{{UCF101 (\%)}}&\multicolumn{2}{c|}{{Kinetics (\%)}} &\multicolumn{2}{c|}{{SSv2 (\%)}}
&\multicolumn{2}{c}{{SSv2-small (\%)}}\\
\rowcolor{mygray} 
\multicolumn{2}{c||}{\multirow{-2}{*}{Method}}   & $1$-shot   & $5$-shot & $1$-shot    & $5$-shot  & $1$-shot   & $5$-shot  & $1$-shot  & $5$-shot & $1$-shot   & $5$-shot \\
\hline \hline

CLIP-FSAR~\cite{wang2023clip}\!\!\!\!&\!\!\pub{IJCV24}         & $77.1$    & $87.7$ &    $97.0$    & $99.1$   & $94.8$    & $95.4$  & $62.1$    & $72.1$ & $54.6$    & $61.8$ \\
\hline \hline
\multicolumn{2}{c||}{\textsc{DiST} \textbf{(Ours)}}                   & $\mathbf{82.6}$    & $\mathbf{88.7}$&  $\mathbf{98.3}$      & $\mathbf{99.2}$  & $\mathbf{95.6}$   & $\mathbf{96.0}$ & $\mathbf{64.6}$  & $\mathbf{75.8}$&  $\mathbf{57.5}$  & $\mathbf{63.5}$ \\
\hline
\end{tabular}}
\vspace{-5pt}
\label{table::ensable}
\end{table*}
\section{Experiments}

\subsection{Experimental Setup}
\sssection{Dataset.} We conduct extensive experiments on five datasets,~\ie, Kinetics~\cite{carreira2017quo}, SSv2-full~\cite{goyal2017something}, SSv2-small~\cite{goyal2017something}, HMDB51~\cite{kuehne2011hmdb}, and UCF101~\cite{soomro2012ucf101}. For SSv2-full~\cite{goyal2017something}, SSv2-Small~\cite{goyal2017something} and Kinetics~\cite{carreira2017quo}, we utilize the split as in CMN~\cite{zhu2018compound}, with $64$, $12$, and $24$ classes used for \texttt{train}, \texttt{val}, and \texttt{test}, respectively. For HMDB51~\cite{kuehne2011hmdb} and UCF101~\cite{soomro2012ucf101}, we adopt the split setting as in ARN~\cite{zhang2020few}, where the $51$ classes in HMDB51 are split into $31$/$10$/$10$ classes for \texttt{train}/\texttt{val}/\texttt{test}, while the $101$ classes in UCF101 are split into $70$/$10$/$21$ classes for \texttt{train}/\texttt{val}/\texttt{test}.

\sssection{Evaluation.} Following the official evaluation $\!$protocols~\cite{boosting,cao2020few}, we use $5$-way $1$-shot  and  $5$-way $5$-shot accuracy for evaluation, and report the average results over 10,000 tasks randomly selected from \texttt{test}.


\subsection{Implementation Details}
\sssection{Network Architecture.} We use CLIP ViT-B~\cite{radford2021learning} as our backbone for a fair comparison with previous methods~\cite{wang2023clip,wang2023few}. Note that the backbone in \textsc{DIST} is initialized with the pre-trained CLIP ViT-B/16 parameters. By default, the number of spatial attributes $G$ and temporal attributes $L$ are set to $6$ and $3$, respectively (ablation study in Table~\ref{tab:promptnum}). The number of object-level prototypes $N$ is $9$. The value of parameter $\alpha$ is set to $0.5$ (see Fig.~\ref{class_gain}~{\color{red}(left)}).

\sssection{Network Training.} Following previous methods~\cite{wang2016temporal,wang2022hybrid,cao2020few,boosting}, we uniformly and sparsely sample $T$ = $8$ frames from each video to encode video representation. In the training phase, we adopt basic data augmentation, such as random horizontal flipping, cropping, and color jitter. In contrast, only a center crop is used during the testing phase. To retain the original pre-trained prior knowledge in the text encoder and reduce the optimization burden, we freeze the text encoder, ensuring it remains unchanged during training. Moreover, we use the Adam~\cite{kingma2014adam} optimizer with the multi-step scheduler to train our framework.

\sssection{Reproducibility.} \textsc{DiST} is implemented in PyTorch, and trained on two NVIDIA Tesla V100 GPUs with a $32$GB memory per card. Full code is released at \href{https://github.com/quhongyu/DiST}{DiST}.

\subsection{Comparison with State-of-the-Arts}
\label{sec43}
We compare $\!$the performance $\!$of $\!$our \textsc{DiST} with  current state-of-the-art FSAR methods on five standard datasets~\cite{kuehne2011hmdb,soomro2012ucf101,goyal2017something,carreira2017quo} in Table~\ref{table1} and Table~\ref{table2}. $\!$It demonstrates that \textsc{DiST} outperforms all FSAR methods. $\!$Specifically, compared to CLIP-FSAR~\cite{wang2023clip} that only uses naive class names as semantic information, our approach achieves better results in multiple datasets and task settings. $\!$It indicates our \textsc{DiST} further boosts performance by grasping spatiotemporal-decoupled prior knowledge from LLM to compensate for visual features. $\!$Further, the performance margin between \textsc{DiST} and CLIP-FSAR is more significant under low shots. $\!$Notably, on HMDB51~\cite{kuehne2011hmdb} and UCF101~\cite{soomro2012ucf101} datasets, $\!$the performance of our \textsc{DiST} on the $5$-way $3$-shot setting is comparable to the performance of CLIP-FSAR on the $5$-way $5$-shot setting. We attribute the more pronounced performance gains in the 1-shot setting to the fact that our framework effectively mitigates the extreme visual scarcity when learning from one single example. Since 1-shot support video is often insufficient to capture the spatio-temporal variance of an action, our LLM-generated decoupled prompts function as robust semantic anchors, providing rich spatio-temporal knowledge. By transforming unseen categories into commonsense descriptions, \textsc{DIST} effectively compensates for the critical visual context that naive category names fail to provide. Conversely, in 5-shot scenarios, the relative impact of this external knowledge is moderated, as multiple visual samples provide adequate visual context that can alleviate category ambiguity. Furthermore, Table~\ref{table::ensable} compares our \textsc{DiST} against CLIP-FSAR under another setting, which makes few-shot predictions with the help of zero-shot results. The results show \textsc{DiST} consistently outperforms CLIP-FSAR on each dataset under any setting.

\begin{table*}[!t]
\caption{\textbf{$_{\!}$A$_{\!}$ set$_{\!}$ of$_{\!}$ ablation$_{\!}$ studies$_{\!}$ on$_{\!}$ HMDB51~\cite{kuehne2011hmdb} and SSv2-small~\cite{goyal2017something}}(see$_{\!}$ \S\ref{sec44}). The$_{\!}$ adopted$_{\!}$ designs$_{\!}$ are$_{\!}$ marked$_{\!}$ in$_{\!}$ {\color{red}red}.}
        \centering
        \small
	\begin{subtable}
  {0.5\linewidth}
 \resizebox{\textwidth}{!}{
			\setlength\tabcolsep{3pt}
			\renewcommand\arraystretch{0.91}
     \begin{tabular}{cc|cc|cc|cc}
\thickhline
\rowcolor{mygray} \multicolumn{2}{c}{\textit{Spatial Attribute}}  & \multicolumn{2}{|c|}{\textit{Temporal Attribute}} & \multicolumn{2}{c|}{HMDB51} & \multicolumn{2}{c}{SSv2-small}\\ 
\rowcolor{mygray} Concat       & SKC 
          &      Concat     & TKC &   $1$-shot     &  $5$-shot  &   $1$-shot     &   $5$-shot \\
\hline \hline
 \cmark      & &  \cmark    &   &  $80.7$   &   $88.3$&   $55.0$   &    $61.6$ \\
     &\cmark &    \cmark     &  &  $81.0$ &  $88.6$  &  $55.9$   &    $62.7$\\
      \cmark     &  &      & \cmark   &   $81.6$ & $88.5$ &   $56.4$   &    $63.0$\\  
     \arrayrulecolor{gray}\hdashline\arrayrulecolor{black}
             & {\color{red}\cmark}  &    & {\color{red}\cmark}   & $\mathbf{82.6}$  & $\mathbf{88.7}$&  $\mathbf{57.5}$ &  $\mathbf{63.4}$ \\
\hline
\end{tabular}
	}
		\setlength{\abovecaptionskip}{0.3cm}
		\setlength{\belowcaptionskip}{-0.1cm}
		\caption{attribute injection manner}
		\vspace{3px}
        \label{prompt_manner}
	\end{subtable}
        \hspace{-0.7em}     
        \begin{subtable}{0.5\linewidth}
         \centering
        \small
		\resizebox{\textwidth}{!}{
			\setlength\tabcolsep{2.5pt}
			\renewcommand\arraystretch{0.94}
        \begin{tabular}{cc|cc|cc|cc}
\thickhline
\rowcolor{mygray}
\multicolumn{2}{c|}{\textit{Spatial  Attribute}}  & \multicolumn{2}{c|}{\textit{Temporal  Attribute}} & \multicolumn{2}{c|}{HMDB51}  & \multicolumn{2}{c}{SSv2-small} \\   
\rowcolor{mygray} Label       &   Knowledge 
          &      Label     &   Knowledge &   $1$-shot     &  $5$-shot &   $1$-shot     &  $5$-shot   \\
\hline \hline
 \cmark    & &    \cmark     &  &     $80.0$ & $87.3$  &     $54.7$ & $61.4$   \\
      \     & \cmark &   \cmark   &    &   $81.2$ & $88.0$  &     $55.7$ & $62.8$  \\  
      \cmark      &  &      & \cmark  &  $81.6$   &   $88.6$  &     $56.5$ & $62.9$ \\
      \arrayrulecolor{gray}\hdashline\arrayrulecolor{black}
               & {\color{red}\cmark}  &    & {\color{red}\cmark}   & $\mathbf{82.6}$  & $\mathbf{88.7}$  & $\mathbf{57.5}$  & $\mathbf{63.4}$ \\
\hline
\end{tabular}
}
		\setlength{\abovecaptionskip}{0.3cm}
		\setlength{\belowcaptionskip}{-0.1cm}
		\caption{attribute content}
		\vspace{3px}
		\label{prompt_content}
	\end{subtable} \\
 \begin{subtable}{0.3\linewidth}
  \vspace{-9px}
  \centering
        \small
		\resizebox{\textwidth}{!}{
			\setlength\tabcolsep{2pt}
			\renewcommand\arraystretch{1.1}
\begin{tabular}{c|ccc}
\thickhline
\rowcolor{mygray}
 Method  & HMDB51 & SSv2-small \\ \hline  \hline
  CLIP-FSAR~\cite{wang2023clip}      & $75.8$ & $53.8$    \\      
  CLIP-FSAR$^{\dagger}$     & $81.0$   & $56.1$     \\ 
   \arrayrulecolor{gray}\hdashline\arrayrulecolor{black}
 \textsc{DiST} \textbf{(Ours)}          & $\mathbf{82.6}$ & $\mathbf{57.5}$     \\ \hline
\end{tabular}	}
		\setlength{\abovecaptionskip}{0.3cm}
		\setlength{\belowcaptionskip}{-0.1cm}
		\caption{knowledge compensator}
		\vspace{-0.6cm}
		\label{tab::knwo_ledge}
	\end{subtable}
        \hspace{-0.7em}
	\begin{subtable}{0.35\linewidth}
  \vspace{-9px}
  \centering
        \small
		\resizebox{\textwidth}{!}{
			\setlength\tabcolsep{3pt}
			\renewcommand\arraystretch{1.12}
       \begin{tabular}{l|cc}
\thickhline
\rowcolor{mygray} Spatial Metric
  & HMDB51 & SSv2-small \\
\hline \hline
One-to-one matching & $82.4$       &$56.6$      \\
Bi-MHM~\cite{wang2022hybrid} &$ 82.4$       & $57.1$      \\
  \arrayrulecolor{gray}\hdashline\arrayrulecolor{black}
\textbf{Ours}                          & $\mathbf{82.6}$        & $\mathbf{57.5}$  \\
\hline
\end{tabular}	}
		\setlength{\abovecaptionskip}{0.3cm}
		\setlength{\belowcaptionskip}{-0.1cm}
		\caption{spatial matching metric}
		\vspace{-0.6cm}
		\label{Spatial_Metric}
	\end{subtable}
        \hspace{-0.7em}
	\begin{subtable}{0.35\linewidth}
 \vspace{-9px}
  \centering
        \small
		\resizebox{\textwidth}{!}{
			\setlength\tabcolsep{3pt}
			\renewcommand\arraystretch{0.85}
  \begin{tabular}{l|cc}
\thickhline
\rowcolor{mygray} Temporal Metric & HMDB51 & SSv2-small  \\ \hline \hline
  
   CLIP-FSAR (Bi-MHM)       & $76.0$ &  $54.0$  \\ 
  \textbf{Ours (Bi-MHM)}      & $\mathbf{82.7}$  & $\mathbf{57.7}$   \\  \hline

   CLIP-FSAR (OTAM)       & $75.8$ &  $53.8$ \\ 
      \textbf{Ours (OTAM)}  & $\mathbf{82.6}$   & $\mathbf{57.5}$   \\
       \hline
\end{tabular}
		}
		\setlength{\abovecaptionskip}{0.3cm}
		\setlength{\belowcaptionskip}{-0.1cm}
		\caption{temporal matching metric}
		\vspace{-0.6cm}
		\label{Temporal_Metric}
	\end{subtable}
\label{tab::abs}
\end{table*}

\begin{table}[!t]
\caption{\textbf{Impacts of core components} on HMDB51~\cite{kuehne2011hmdb}, SSv2-small~\cite{carreira2017quo}, and UCF101~\cite{soomro2012ucf101}
under  $1$-shot tasks (\S\ref{sec44}).}
\centering
\small
\resizebox{0.49\textwidth}{!}{
\setlength\tabcolsep{2pt}
\renewcommand\arraystretch{1.1}
\begin{tabular}{c|ccc}
\thickhline
\rowcolor{mygray}
 Method Component & HMDB51(\%) & SSv2-small(\%) & UCF101(\%) \\ \hline  \hline
  \textsc{Baseline}        & $75.8$ & $53.8$  & $96.0$   \\  \arrayrulecolor{gray}\hdashline\arrayrulecolor{black}
       SKC \textit{only}        & $77.6$ & $55.7$ & $96.6$   \\ 
   TKC \textit{only}      & $81.0$   & $56.1$ & $97.9$     \\ 
   \arrayrulecolor{gray}\hdashline\arrayrulecolor{black}
 \textsc{DiST} \textbf{(Ours)}          & $\mathbf{82.6}$ & $\mathbf{57.5}$   & $\mathbf{98.3}$   \\ \hline
\end{tabular}} 
\vspace{-8pt}
\label{core_module}
\end{table}
\subsection{Ablation Study}
\label{sec44}
We conduct comprehensive ablation experiments to evaluate the efficacy of our idea and core model designs. Unless otherwise specified, all experiments are conducted adopting the CLIP-ViT-B model as the default backbone.

\subsubsection{Key Component Analysis} Table~\ref{core_module}  summarizes the impact of each module in \textsc{DiST}. 
Here \textsc{BASELINE} denotes \textsc{DIST} only relies on category names as semantic features and performs simple temporal modeling on frame-level features and these semantic features  without TKC and SKC as in CLIP-FSAR~\cite{wang2023clip}.
Specifically, compared to the baseline, \textbf{Temporal Knowledge Compensator (TKC)} (\cf \S\ref{TKC}) brings $\mathbf{5.2}$\%, $\mathbf{2.3}$\% and $\mathbf{1.9}$\% performance gains on HMDB51~\cite{kuehne2011hmdb}, SSv2-small~\cite{carreira2017quo} and UCF101~\cite{soomro2012ucf101}, respectively. This consistent promotion indicates that TKC can enhance the temporal awareness of \textsc{DiST} to facilitate accurate matching. In addition, the proposed \textbf{Spatial Knowledge Compensator (SKC)} (\cf \S\ref{SKC}) improves on the three datasets~\cite{kuehne2011hmdb,soomro2012ucf101} by $\mathbf{1.8}$\%, $\mathbf{1.9}$\% and $\mathbf{0.6}$\%, respectively, which indicates leveraging spatial prior knowledge can focus on action-related spatial details to boost few-shot performance. Moreover, combining the two modules can further improve performance, indicating the complementarity between two core modules.

\begin{figure*}[!t]
\centering
\begin{subfigure}{0.4\textwidth}
\includegraphics[width=0.83\textwidth]{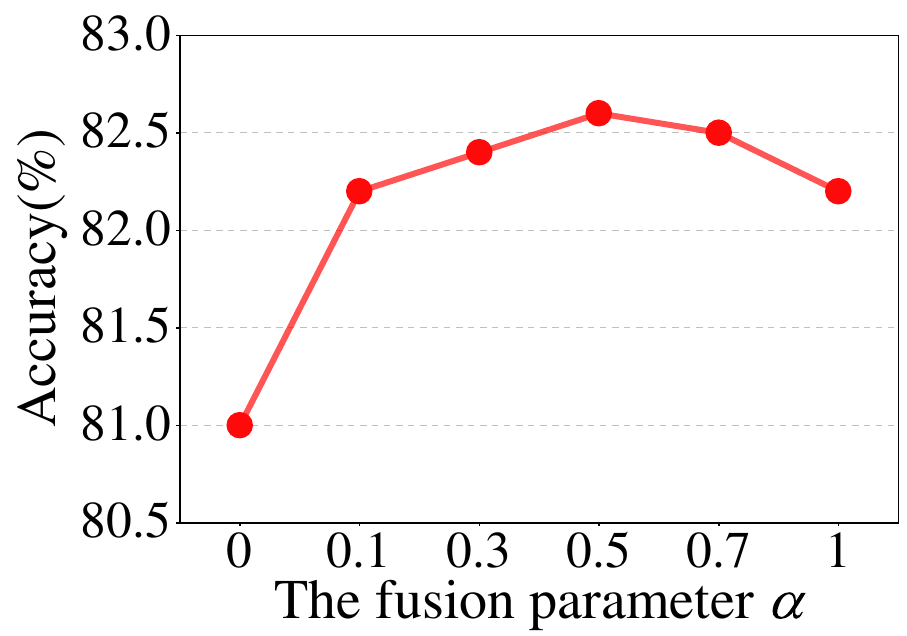}
\end{subfigure}%
\hspace{-0.7em}
\begin{subfigure}{0.6\textwidth}
  \flushright
  \includegraphics[width=0.98\textwidth]{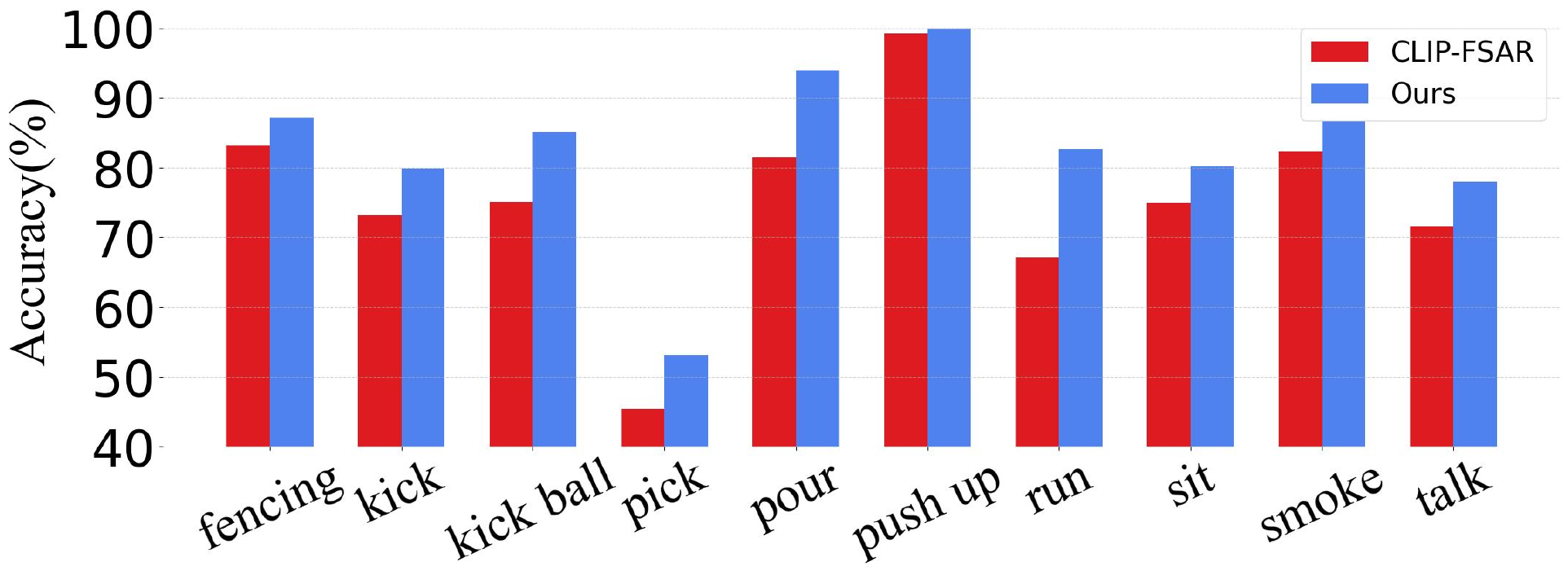}
\end{subfigure}
\vspace{-3pt}
\caption{\textbf{Left}: \textbf{The impact of the varying fusion parameter} $\alpha$  on HMDB51~\cite{kuehne2011hmdb} in the $5$-way $1$-shot setting (see \S\ref{sec44}). \textbf{Right}: \textbf{5-way 1-shot class improvement of \textsc{DiST}} compared to CLIP-FSAR~\cite{wang2023clip} on all class action classes on HMDB51~\cite{kuehne2011hmdb} (see \S\ref{sec45}). Our \textsc{DiST} achieves improvement on all action classes.}
\label{class_gain}
\vspace{-7pt}
\end{figure*}

\subsubsection{Impact of Different Numbers of Prompts}\label{secB1}
We next investigate the impact of various prompt configurations in Table~\ref{tab:promptnum}. First, we conduct experiments with $1$, $3$, $6$, $9$ and $12$ spatial attributes. We can observe that the performance reaches its peak at $G\!=\!6$,  possibly because too many spatial attributes may introduce redundant or noisy local cues, while too few spatial attributes may fail to capture sufficient spatial context for effective few-shot action recognition. Furthermore, the best result is obtained when the number of temporal attributes is $L\!=\!3$. We speculate that too few temporal attributes may fail to adequately convey the temporal changes in actions while too many temporal attributes often contain noisy temporal information, which leads to performance degradation. Therefore, we set $\{G\!=\!6, L\!=3\!\}$ as the default configuration to strike an optimal balance between accuracy and computation cost.

\begin{table}[!t]
\centering
\caption{The ablation study on HMDB51~\cite{kuehne2011hmdb} to investigate the
configuration of the number of spatial attributes $G$ and temporal attributes $L$ (see \S\ref{sec44}).}
\label{tab:promptnum}
\resizebox{0.38\textwidth}{!}{
\setlength\tabcolsep{15pt}
\renewcommand\arraystretch{0.95}
\begin{tabular}{l|cc}
\thickhline
\rowcolor{mygray}
 & 
 \multicolumn{2}{c}{{HMDB51 (\%)}}\\
\rowcolor{mygray} \multirow{-2}{*}{\{$G$,$L$\}}
& $1$-shot      & $5$-shot \\
\hline\hline
$\{1, 3\}$ & $81.8$       & $87.9$      \\
$\{3, 3\}$ & $82.4$       & $88.5$      \\
$\{9, 3\}$ & $82.5$       & $88.5$      \\
$\{12, 3\}$ &$82.3$       & $88.3$      \\ \hline
$\{6, 1\}$ &$81.4$       & $88.0$      \\
$\{6, 2\}$ &$81.9$       & $88.2$      \\
$\{6, 4\}$ &$82.0$       & $88.5$      \\
$\{6, 5\}$ &$81.8$       & $88.3$      \\
$\{6, 6\}$ &$ 81.5$       & $88.0$      \\ \hline
$\{6, 3\}$                           & $\mathbf{82.6}$        & $\mathbf{88.7}$  \\
\hline
\end{tabular}
}
\vspace{-8pt}
\end{table}

\subsubsection{Impact of Different Attribute Injection Manners} We respectively propose SKC (\cf \S\ref{SKC}) and TKC (\cf \S\ref{TKC}) to inject spatial/temporal attribute features into visual features. In Table~\ref{tab::abs}a, we study the effect of different temporal/spatial attribute injection manners. ``Concat" means that the visual features and attribute features are directly concatenated and then fed into transformers for multimodal fusion like CLIP-FSAR~\cite{wang2023clip}. The experimental results show that our proposed SKC and TKC yield better results,  suggesting the effectiveness of our module design for FSAR.

\subsubsection{Impact of Different Attribute  Content}
We investigate the impact of different temporal and spatial attribute content on the performance of our proposed \textsc{DiST} in Table~\ref{tab::abs}b.  \textit{Different} from CLIP-FSAR~\cite{wang2023clip} only using class labels and visual features for simple fusion, the baseline model in the first row feeds the class labels and patch-level/frame-level visual features into Spatial/Temporal Knowledge Compensators (SKC/TKC) to discover discriminative object-level and frame-level prototypes for a fair comparison with other models in the table. Note that the network architecture (\emph{i.e.}, SKC and TKC) of different model variants in Table 4(b) remains constant, and the only change is the source of the semantic information (\emph{i.e.}, using class names versus decomposed knowledge) injected into  SKC and TKC. This setup allows us to position the contribution of LLM-generated spatial and temporal attributes relative to naive category labels.
As seen, we observe that utilizing spatial and temporal prior knowledge generated by LLM consistently performs better than using class names, with  $\mathbf{1.2}$\% and $\mathbf{1.6}$\% performance gains in the $1$-shot setting on HMDB~\cite{kuehne2011hmdb}, respectively. In addition, combining both spatial and temporal prior knowledge yields better results, which demonstrate that the integration of different prior knowledge is complementary. 

\begin{table}[!t]
\centering
\caption{The ablation study on HMDB51~\cite{kuehne2011hmdb} and SSv2-small~\cite{goyal2017something} under the 1-shot setting to investigate the number of object-level prototypes $N$ (see \S\ref{sec44}).}
\label{tab:pronum}
\resizebox{0.45\textwidth}{!}{
\setlength\tabcolsep{7pt}
\renewcommand\arraystretch{1.0}
\begin{tabular}{c|cc}
\thickhline
\rowcolor{mygray} Prototype Number $N$
& HMDB51     & SSv2-small\\
\hline\hline
$N=1$&   $80.8$   &    $56.4$  \\
$N=4$ &    $81.7$   &   $56.9$    \\
$N=\mathbf{9}$ &  $\mathbf{82.6}$     &  $\mathbf{57.5}$  \\

$N=12$ &$82.3$      &$57.2$     \\ 
$N=16$ & $82.1$        & $57.0$   \\
\hline
\end{tabular}
}
\vspace{-8pt}
\end{table}
\subsubsection{Impact of Knowledge Compensators} 
To better position the contribution of our knowledge compensators, we construct a variant of CLIP-FSAR~\cite{wang2023clip} by simply replacing its original class labels with our LLM-generated prompts (\ie, CLIP-FSAR$^{\dagger}$). The comparison results are reported in Table~\ref{tab::abs}c. As seen, \textsc{DiST} gains larger improvements compared to CLIP-FSAR$^{\dagger}$. This suggests our performance gains are not solely due to the usage of LLM-generated prompts, but also due to proposed knowledge compensators which explicitly inject spatial attributes into patch features and temporal attributes into frame features rather than using simple global fusion. Thus, these compensators construct more informative prototypes that facilitate FSAR. The substantial performance gap between CLIP-FSAR$^{\dagger}$ and \textsc{DiST}  underscores that the synergistic combination of high-quality knowledge and a dedicated injection architecture is essential for learning informative prototypes that facilitate few-shot action recognition.

\subsubsection{Impact of the Number of Object-level Prototypes}
We next investigate the impact of the number $N$ of object-level prototypes in Spatial Knowledge Compensator, which are designed to aggregate image content and highlight potential target objects from patch tokens. As summarized in Table \ref{tab:pronum}, we vary $N$ from 1 to 16 on HMDB51~\cite{kuehne2011hmdb} and SSv2-small~\cite{goyal2017something}. 
We can clearly observe that, our algorithm gains stable improvements (\ie, $80.8\!\rightarrow\!82.6$ on HMDB51) as the number of prototypes grows (\ie, $N\!=\!9$). This supports our hypothesis that maintaining multiple object-level prototypes allows the model to better capture diverse action-related entities and fine-grained spatial contexts.
However, further increasing $N$ above 9 results in a slight performance drop. We speculate that too many prototypes may introduce noisy background information during the patch aggregation process.
Thus, we empirically set $N\!=\!9$ as the default configuration to ensure an optimal balance between recognition accuracy and computational efficiency.

 \begin{table*}[!t]
\centering
    \caption{\textbf{Complexity analysis} for $5$-way $1$-shot HMDB51~\cite{kuehne2011hmdb} and Kinetics~\cite{carreira2017quo} evaluation. Here, we report Params, FLOPs, GPU memory, and Speed for each model.  See \S\ref{sec44} for more details. }
 \resizebox{0.85\textwidth}{!}{  
 \setlength\tabcolsep{8pt}
    \renewcommand\arraystretch{1}
 \begin{tabular}{c|c|c|c|c|c|c}
\thickhline
\rowcolor{mygray} Method  & Params & FLOPs & Memory & Speed & HMDB51 (\%) &Kinetics (\%) \\
 \hline\hline
 CLIP-FSAR~\cite{wang2023clip}          & $89.3$M &  $901.9$G &$13.8$G&$36.7$ms & $75.8$ & $89.7$  \\ 
 \textsc{DiST}(ours)   & $97.2$M &  $902.1$G &$14.1$G &$40.9$ms  &  $82.6$  & $92.7$   \\ \hline 
\end{tabular}}
    \label{efficiency}
    \vspace{-10pt}
\end{table*}

\begin{table}[!t]
\caption{\textbf{Comparison results with different large language models used for spatial-temporal attribute generation} on HMDB51~\cite{kuehne2011hmdb} and UCF101~\cite{soomro2012ucf101} in the $5$-way$_{\!}$ $1$-shot$_{\!}$ tasks$_{\!}$ (see$_{\!}$ \S\ref{sec44}). \textsc{Baseline} denotes that  DIST only relies
on category names for TKC and SKC.}
\centering
\small
\resizebox{0.42\textwidth}{!}{
\setlength\tabcolsep{7pt}
\renewcommand\arraystretch{1.1}
\begin{tabular}{c|cc}
\thickhline
\rowcolor{mygray}
 LLM version & HMDB51 (\%) & UCF101 (\%)\\ \hline  \hline
  \textsc{Baseline}        & $75.8$ & $96.0$     \\  
       Llama-2~\cite{touvron2023llama}& $82.1$ & $98.2$   \\ 
   Vicuna~\cite{chiang2023vicuna}      & $82.5$   & $98.0$      \\ 
   \arrayrulecolor{gray}\hdashline\arrayrulecolor{black}
   GPT-3.5~\cite{ChatGPT} (ours)          & $\mathbf{82.6}$ & $\mathbf{98.3}$      \\ \hline
\end{tabular}} 
\vspace{-4pt}
\label{tab::llm}
\end{table}
\begin{table}[!t]
\caption{\textbf{Impact of fine-tuning visual encoder} on the generalization capacity of \textsc{DIST} on HMDB51~\cite{kuehne2011hmdb}, Kinetics~\cite{carreira2017quo}, and SSv2-small~\cite{goyal2017something}
in the $5$-way $1$-shot tasks (see$_{\!}$ \S\ref{sec44}).}
\centering
  \small
  \resizebox{0.49\textwidth}{!}{
    \setlength\tabcolsep{1pt}
    \renewcommand\arraystretch{1}
    \begin{tabular}{r|c|cccc}
      \thickhline
     \rowcolor{mygray} {Method} &{Visual encoder} & {HMDB51} & {Kinetics} & {SSv2-small}   \\
      \hline\hline
      CLIP-Freeze ~\cite{pei2023d}& Frozen & 58.2 &78.9 & 29.5 \\
      \textsc{DiST} (ours)&Frozen & 62.6  &83.1 &35.9   \\\hline
      \textbf{\textsc{DiST} (ours)} &Fine-tuning & \textbf{84.7}  & \textbf{95.9} &\textbf{60.7}  \\
      \hline
    \end{tabular}}
    \label{tab::finetuning}
    \vspace{-4pt}
\end{table}
\subsubsection{Model Efficiency Analysis}
To analyze the effectiveness of training and inference, we provide a detailed comparison with the state-of-the-art CLIP-FSAR~\cite{wang2023clip} in terms of parameters, FLOPs, GPU memory, and inference speed, as summarized in Table~\ref{efficiency}. For a fair comparison, we adopt the same visual encoder (ViT-B) for both methods. As shown in Table~\ref{efficiency}, compared to SOTA CLIP-FSAR, our additional overhead is minimal. Specifically, the additional FLOPs and memory usage remain within a negligible range compared to CLIP-FSAR. Despite this lightweight design, our \textsc{DiST} can bring $6.8$\% and $3.0$\% accuracy improvements on HMDB51 and Kinetics over CLIP-FSAR, respectively. These results demonstrate that our spatiotemporal knowledge compensators are both effective and efficient, enabling substantial performance improvement without sacrificing computational feasibility.

\subsubsection{Effect of the Fusion Parameter $\alpha$}
We study the sensitivity of our model to the fusion parameter $\alpha$ in Eq.\ref{gongshi2}, which controls the trade-off between spatial and temporal matching scores. Fig.\ref{class_gain}\textcolor{red}{(left)} reports the performance on the HMDB51 dataset~\cite{kuehne2011hmdb} under the 5-way 1-shot setting as $\alpha$ varies in $[0, 1]$.
We observe a clear peak at $\alpha\!=\! 0.5$, indicating that equal weighting of spatial and temporal information yields the best performance. This suggests that spatial and temporal knowledge contribute complementary cues for action recognition under the few-shot setting. Notably, setting $\alpha\!=\!0$ or $\alpha\!=\!1$, which corresponds to relying solely on spatial or temporal information, respectively, leads to a large performance drop.  These results highlight the necessity of jointly modeling fine-grained spatial patterns and dynamic temporal patterns to improve generalization from a few examples.

\subsubsection{Impact of Different Large Language Models for Attribute Generation}
To investigate the effect of different language models as a knowledge base for spatial-temporal attribute generation, we conduct comparisons among ChatGPT (GPT-3.5), LLaMA-2, and Vicuna under identical prompt templates and experimental settings. As shown in Table~\ref{tab::llm}, all variants leveraging large language models consistently outperform the baseline, which only uses raw class names for spatial-temporal feature modeling. This highlights the importance of diverse spatio-temporal attribute knowledge in enhancing FSAR performance. While using GPT-3.5 as a knowledge base yields the highest performance due to its superior reasoning ability, the minor gaps compared to LLaMA-2 and Vicuna show the model-agnostic nature of our framework, enabling broad applicability across different LLMs.

\begin{table}[!t]
\caption{\textbf{Comparison results between Spatial Knowledge Compensator and existing spatial interaction methods} on HMDB51~\cite{kuehne2011hmdb}, SSv2-small~\cite{carreira2017quo}, and UCF101~\cite{soomro2012ucf101}
under  $1$-shot tasks (\S\ref{sec44}).}
\centering
\small
\resizebox{0.49\textwidth}{!}{
\setlength\tabcolsep{2pt}
\renewcommand\arraystretch{1.1}
\begin{tabular}{c|ccc}
\thickhline
\rowcolor{mygray}
 Spatial Interaction & HMDB51(\%) & SSv2-small(\%) & UCF101(\%) \\ \hline  \hline
\textsc{Baseline}        & $75.8$ &$53.8$  & $96.0$   \\  \arrayrulecolor{gray}\hdashline\arrayrulecolor{black}
       PLE~\cite{thatipelli2022spatio}        & $76.0$ & $54.1$ & $95.8$  \\ 
  STMM~\cite{xing2023revisiting}     & $76.2$   & $54.4$ &$96.1$     \\ 
   \arrayrulecolor{gray}\hdashline\arrayrulecolor{black}
 \textsc{DiST} \textbf{(Ours)}         & $\mathbf{77.6}$ & $\mathbf{55.7}$  & $\mathbf{96.6}$   \\ \hline
\end{tabular}} 
\vspace{-8pt}
\label{tab::skc}
\end{table}

\subsubsection{Impact of Fine-tuning Visual Encoder}
To further investigate the influence of fine-tuning the visual encoder on the generalization capacity of \textsc{DIST}, we compare \textsc{DIST} against our model freezing the visual encoder across HMDB51~\cite{kuehne2011hmdb}, Kinetics~\cite{carreira2017quo}, and SSv2-small~\cite{goyal2017something}. As shown in Table \ref{tab::finetuning}, while maintaining a frozen visual encoder provides a competitive baseline, fine-tuning the visual backbone leads to a substantial performance leap. Specifically, \textsc{DIST} achieves a remarkable 84.7\% on HMDB51, 95.9\% on Kinetics, and 60.7\% on SSv2-small under the $5$-way $1$-shot setting. This significant improvement demonstrates that task-specific fine-tuning can effectively unleash the rich semantic knowledge embedded in CLIP and adapt it to the complex temporal dynamics of action recognition. These results confirm that the architectural innovation of our Knowledge Compensators, coupled with backbone adaptation, is essential for achieving state-of-the-art results in the challenging few-shot action recognition.

\subsubsection{Comparison with Spatial Interaction Methods}
To further validate the superiority of our Spatial Knowledge Compensator (SKC), we compare it against representative spatial interaction methods (\ie,  patch-level enrichment (PLE)~\cite{thatipelli2022spatio} and Short-Term Temporal Modeling (STMM)~\cite{xing2023revisiting} ), which employ dense patch-level interactions within each frame and across frames, respectively. We replace SKC with these spatial interaction methods, and the comparison results are summarized in Table~\ref{tab::skc}. As seen, our SKC consistently surpasses these spatial interaction methods across HMDB51~\cite{kuehne2011hmdb}, SSv2-small~\cite{carreira2017quo}, and UCF101~\cite{soomro2012ucf101} datasets under the $1$-shot setting. Notably, compared to STMM, SKC achieves a noticeable performance boost of 1.4\% on HMDB51 and 1.3\% on SSv2-small. This study confirms that although previous dense interactions capture fine-grained spatial dependencies, they suffer from background noise and a lack of semantic focus. In contrast, our SKC utilizes LLM-generated spatial attributes to guide the aggregation of patch tokens into compact object-level prototypes. Thus, this sparse aggregation strategy effectively filters out irrelevant background context and focuses on class-relevant entities.
\begin{figure*}[htb]
	\centering
	\subfloat[]{\includegraphics[width=0.98\linewidth]{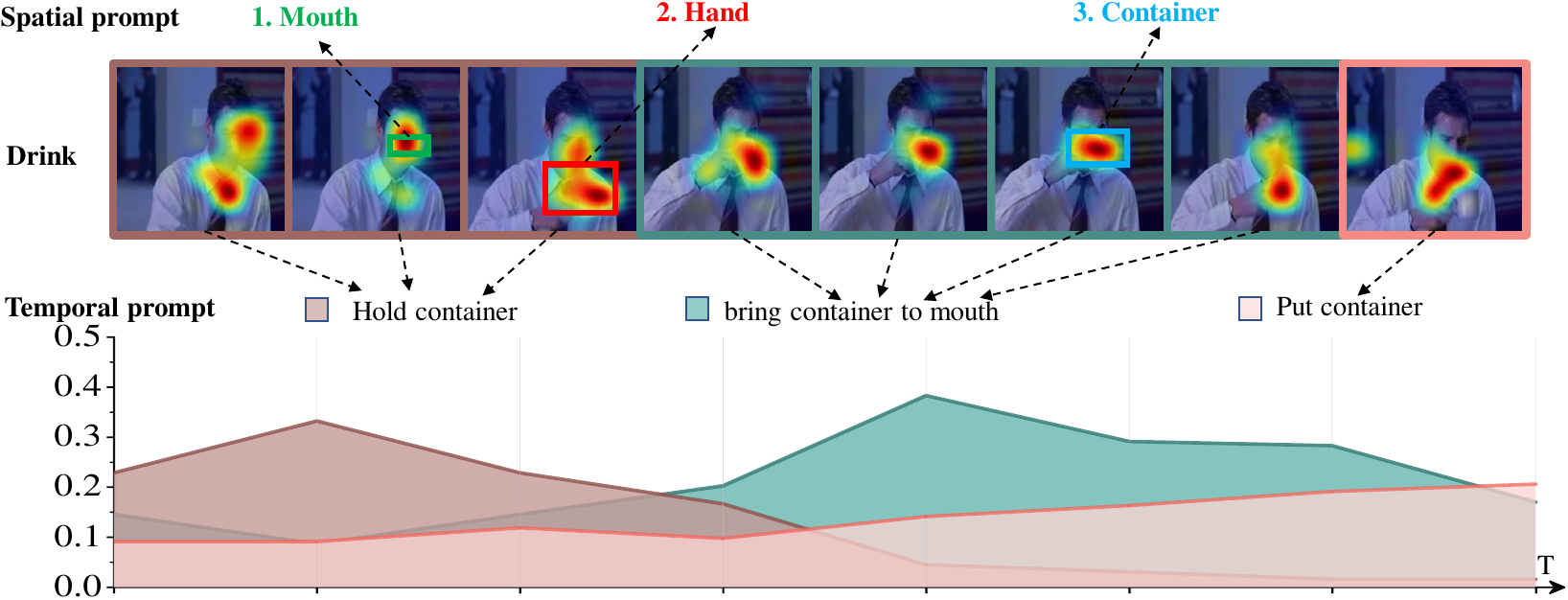}}
	\hfil
	\subfloat[]{\includegraphics[width=0.99\linewidth]{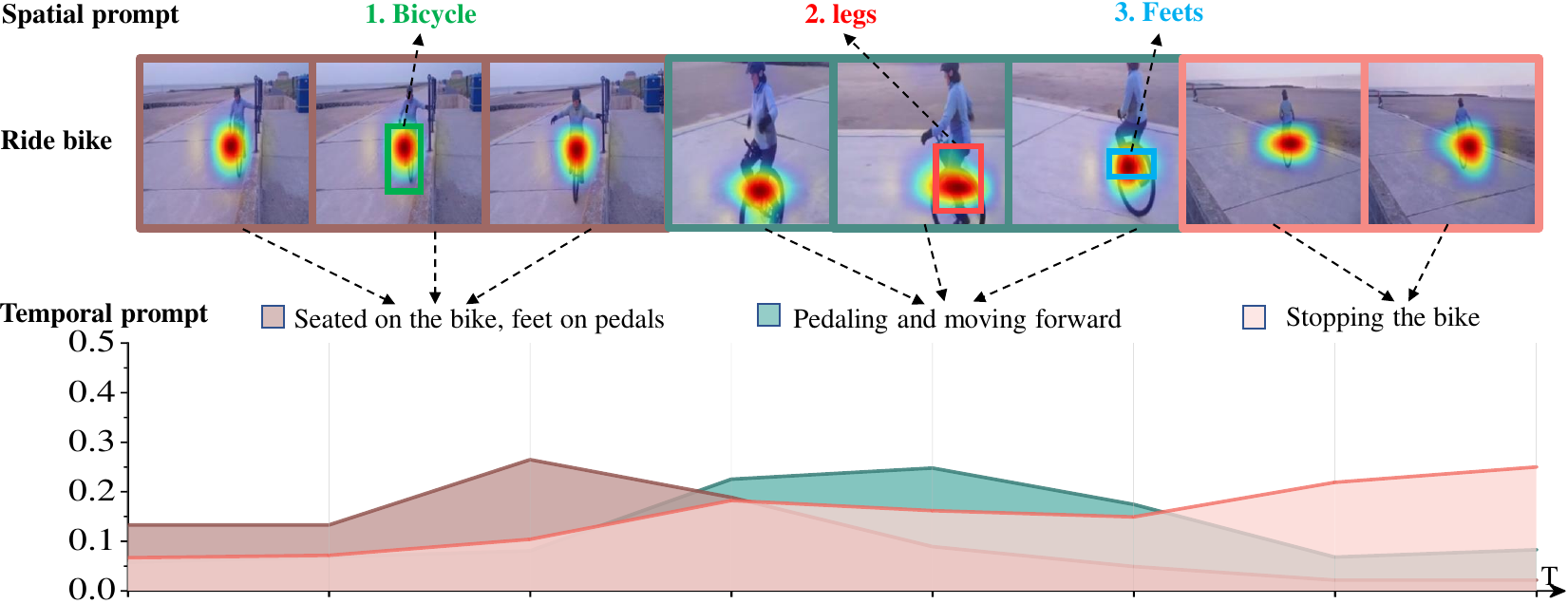}}
	\caption{\textbf{Visualization of spatial and temporal prompts} under the $1$-shot setting (\S\ref{sec45}). The spatial prompts are shown as highlighted response areas in each frame. We also show cross-attention temporal prompt weights of Eq.~\ref{gongshi} in a line graph.}
     \label{att_map}
     \vspace{-5pt}
\end{figure*}

\subsection{Generalization Study}
\label{genstu}
\subsubsection{Matching Metrics}
We analyze the impact of different spatial matching metrics in Table~\ref{tab::abs}d. $\!$We adopt different spatial matching metrics (\cf~Eq.~\ref{eq::spatial_metric}), $\!$including one-to-one matching, Bi-MHM~\cite{wang2022hybrid}, $\!$and our proposed spatial metric. $\!$One-to-one matching means computing the spatial matching scores of aligned object-level prototypes between the support video and query video. $\!$The results show that our proposed spatial metric achieves the best results, suggesting the effectiveness of our proposed spatial matching metric. We also conduct experiments using different temporal matching metrics (\cf~Eq.~\ref{eq::temporal_metric}) on HMDB51~\cite{kuehne2011hmdb} and SSv2-small~\cite{goyal2017something}. As shown in Table~\ref{tab::abs}e, our method can adapt to any temporal alignment metric and achieves better performance compared to CLIP-FSAR~\cite{wang2023clip}.
\subsubsection{Generalization with ImageNet pre-trained Backbones}
To assess the generalization ability of our  \textsc{DiST} beyond CLIP-based visual encoders~\cite{radford2021learning}, we conduct additional experiments by replacing the default CLIP backbone with two commonly used ImageNet-pretrained architectures: ResNet-50 and ResNet-18, which are single-modal pre-trained initialization. Note that we keep the text encoder unchanged. Table~\ref{tab::backbone} reports comparison results with CLIP-FSAR~\cite{wang2023clip} on various ImageNet pre-trained Backbones. As seen, despite lacking the rich multi-modal pretraining of CLIP, our \textsc{DiST} consistently achieves better performance on any backbones, \eg, outperforming CLIP-FSAR using ResNet-50 by 2.4\% on SSv2-small under the 1-shot setting. This demonstrates the robustness and generalization of our method. Notably, \textsc{DiST} with ResNet-50 outperforms ResNet-18 consistently, aligning with the capacity gap between the two backbones. These results suggest that the core design of \textsc{DiST} is transferable and not restricted to a particular encoder architecture.
\begin{figure*}[!t]
    \centering
\includegraphics[width=0.99\textwidth]{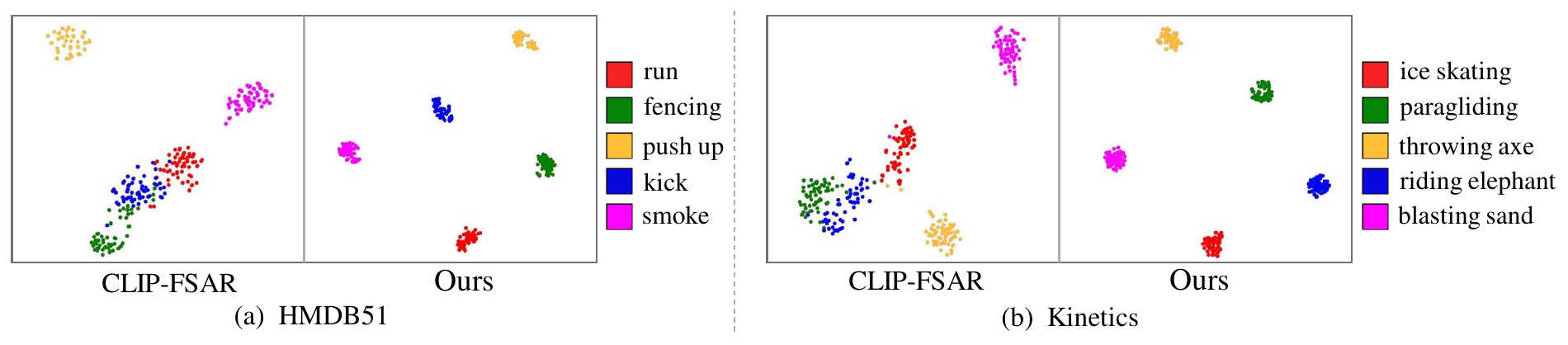}
\vspace{-6pt}
    \caption{\textbf{T-SNE visualization}~\cite{van2008visualizing} under the $5$-way setting on HMDB51 and Kinetics (see \S\ref{sec45}). }
    \vspace{-11pt}
    \label{tsne}
\end{figure*}
\begin{table}[!t]
\caption{\textbf{Generalization performance of our \textsc{DiST} with different ImageNet-pretrained backbones} on SSv2-small~\cite{carreira2017quo}, and Kinetics~\cite{soomro2012ucf101}
in the $5$-way $1$-shot and $5$-shot  tasks (see$_{\!}$ \S\ref{genstu}).}
\centering
\small
\resizebox{0.48\textwidth}{!}{
\setlength\tabcolsep{3pt}
    \renewcommand\arraystretch{1}
    \begin{tabular}{r|c|cc|cc}
      \hline 
     \rowcolor{mygray} &&\multicolumn{2}{c|}{SSv2-small (\%)}&\multicolumn{2}{c}{Kinetics (\%)} \\
     \rowcolor{mygray} \multirow{-2}*{Method}   & \multirow{-2}*{Backbone}  &$1$-shot         & $5$-shot   & $1$-shot       & $5$-shot \\
      \hline\hline
      CLIP-FSAR~\cite{wang2023clip} &  & 46.0 & 52.4 & 72.9&84.1 \\
      \textbf{\textsc{DiST} (ours)}  & \multirow{-2}*{INet-RN18}  & \textbf{48.6} & \textbf{53.3} & \textbf{75.6} & \textbf{86.5}\\
      \hline
      CLIP-FSAR~\cite{wang2023clip} &  & 46.7 & 53.3 & 75.5 & 85.9\\
      \textbf{\textsc{DiST} (ours)}  & \multirow{-2}*{INet-RN50}  & \textbf{49.1} & \textbf{54.6} & \textbf{78.6} & \textbf{88.0}\\
      \hline
      \end{tabular}} 
\label{tab::backbone}
\vspace{-8pt}
\end{table}
\subsubsection{Generalization Under the Parameter-efficient Fine-tuning Setting}
    We further conduct ablation experiments to explore the generalization of our \textsc{DiST} under the parameter-efficient fine-tuning (PEFT) setting. To ensure a fair comparison, we reimplement \textsc{DiST} under the same PEFT setup as recent SOTA methods~\cite{pei2023d,xing2023ma,cao2024task}. As shown in Table~\ref{tab::peft}, \textsc{DiST} consistently surpasses existing PEFT-based approaches under the 1-shot setting, even when using a similar backbone and fine-tuning protocol. This performance gain highlights the strong generalization capability of Spatial and Temporal Knowledge Compensators, which effectively complements visual features to object-level and frame-level prototypes.  The results also confirm that even with limited trainable parameters, \textsc{DiST} effectively transfers knowledge from base to novel classes, demonstrating robustness in FSAR.

\subsection{Quality Analysis}
\label{sec45}
\subsubsection{Class-wise Performance Gains}  $\!$Fig.~\ref{class_gain}~{\color{red}(right)} $\!$shows $5$-way $1$-shot $\!$class-wise $\!$performance $\!$gains $\!$obtained by our \textsc{DiST} over CLIP-FSAR~\cite{wang2023clip} on HMDB51~\cite{kuehne2011hmdb}. Notably, \textsc{DiST} outperforms CLIP-FSAR across all action classes, demonstrating the robustness and generalizability of our approach in the few-shot regime. The most significant improvements, \ie, exceeding 10\% absolute gain, are observed in action categories such as \textit{run}, \textit{pour}, and \textit{kick ball}. These actions typically involve motion patterns and distinctive object interactions, suggesting that our decoupled spatiotemporal modeling effectively captures dynamic visual cues and meaningful spatial contexts. 

\begin{table}[!t]
\caption{\textbf{Generalization performance of our \textsc{DiST} under the parameter-efficient fine-tuning setting} on UCF101~\cite{soomro2012ucf101}, HMDB51~\cite{kuehne2011hmdb}, Kinetics~\cite{carreira2017quo}, SSv2~\cite{goyal2017something} and SSv2-small~\cite{goyal2017something}
in the $5$-way $1$-shot tasks (see$_{\!}$ \S\ref{genstu}).}
\centering
  \small
  \resizebox{0.49\textwidth}{!}{
    \setlength\tabcolsep{1pt}
    \renewcommand\arraystretch{1}
    \begin{tabular}{r|c|cccc}
      \hline
     \rowcolor{mygray} {Method} &{Fine-tuning} & {HMDB51} & {UCF101} & {Kinetics} & {SSv2-small}   \\
      \hline\hline
      D$^{2}$ST-Adapter~\cite{pei2023d}& PEFT & 77.1 & 96.4 &89.3 & 55.0 \\
      MA-FSAR~\cite{xing2023ma}& PEFT & 83.4 & 97.2 & 95.7& 59.1 \\
      Task-Adapter~\cite{cao2024task}&PEFT & 83.6 & 98.0 &95.0 &60.2   \\\hline
      Ours &PEFT & \textbf{84.7}  & \textbf{98.3} & \textbf{95.9} & \textbf{60.7}  \\
      \hline
    \end{tabular}}
    \label{tab::peft}
    \vspace{-8pt}
\end{table}
\subsubsection{Visualization of Feature Distribution} To further qualitatively analyze the changes in feature distribution after incorporating spatiotemporal-decoupled prior knowledge, we follow prior works~\cite{wang2023clip,thatipelli2022spatio} and visualize the feature embeddings using t-SNE for both the baseline CLIP-FSAR~\cite{wang2023clip} (which relies solely on raw category names) and our proposed \textsc{DiST}, as illustrated in Fig.~\ref{tsne}. We observe that after utilizing action-related prior knowledge, our method shows more compact intra-class feature distributions and more discriminative inter-class features. $\!$As shown in Fig.~\ref{tsne}~{\color{red}(a)}, the three classes ``{\color{red} run}", ``{\color{green} fencing}", and ``{\color{blue} kick}" become clearly distinguishable from each other after injecting decoupled attributes into visual features. This demonstrates that the model is better able to distinguish between semantically similar actions by attending to both object-level spatial cues and frame-level temporal dynamics. The enhanced cluster separation reflects the model's capacity to integrate external prior knowledge into visual encoding, thereby boosting generalization in FSAR.

\subsubsection{Visualization of Spatial and Temporal attributes} 
To analyze the role of spatial and temporal prompts in our \textsc{DiST}, we conduct a qualitative study in the $5$-way $1$-shot setting. Fig.~\ref{att_map} displays the visualization results. The attention maps of our \textsc{DiST}  focus more on action-related objects and reduce attention to the background and unrelated objects. This demonstrates our \textsc{DiST} grasps prior knowledge provided by spatial attributes to capture spatial details. Then, we calculate the cross-attention scores between temporal attributes and frames according to Eq.~\ref{gongshi}. It can be seen that different temporal attributes have different weights on the frame sequences, which proves that our \textsc{DiST} can learn temporal relations and capture dynamic semantics. For example, the temporal attribute "Hold container" has larger weights on the first three frames, which indicates these frames may correspond to dynamic semantics implied by the temporal attribute. 

\section{Conclusion}
In this work, we propose \textsc{DiST}, a novel yet effective framework for FSAR that is the first work to grasp spatiotemporal-decoupled prior knowledge from LLM to compensate for visual features. In particular, we design Spatial/Temporal Knowledge Compensators to learn object- and frame-level prototypes, so as to capture fine-grained spatial details and dynamic semantics. Experimental results demonstrate that our \textsc{DiST} achieves state-of-the-art performance on four standard benchmarks. While this work marks an initial step toward incorporating decoupled spatiotemporal priors into FSAR, it paves the way for promising future directions. We believe that further investigation about integrating richer, more structured knowledge from LLMs holds great potential for advancing generalization in low-shot video understanding. 

{
\bibliographystyle{IEEEtran}
\bibliography{main}
}

\end{document}